\documentclass[lettersize,journal]{IEEEtran}
\usepackage{amssymb}
\usepackage{algorithm}
\usepackage{bibentry}
\usepackage{subfigure}
\usepackage{multirow}
\usepackage{epsfig}
\usepackage{color}
\usepackage{epstopdf}
\usepackage{rotating}
\usepackage{caption}
\usepackage{float}
\usepackage{multicol}
\usepackage{ulem}
\usepackage{amsmath,amssymb,amsthm}
\usepackage{bbding}
\usepackage{balance}
\usepackage{microtype}
\usepackage{makecell}
\usepackage{mathtools}
\usepackage{etoolbox}
\usepackage{cases}
\usepackage{enumitem}
\usepackage{xcolor}

\usepackage{tabularx}
\usepackage{threeparttable}
\usepackage{soul}

\usepackage{amsmath,amsfonts}
\usepackage{algorithmic}
\usepackage{array}
\usepackage[caption=false,font=normalsize,labelfont=sf,textfont=sf]{subfig}
\usepackage{textcomp}
\usepackage{stfloats}
\usepackage{url}
\usepackage{verbatim}
\usepackage{graphicx}
\usepackage{booktabs}
\usepackage{color}
\usepackage[most]{tcolorbox}
\tcbuselibrary{theorems}

\def\BibTeX{{\rm B\kern-.05em{\sc i\kern-.025em b}\kern-.08em
    T\kern-.1667em\lower.7ex\hbox{E}\kern-.125emX}}

\begin{document}
\title{MolReFlect: Towards In-Context Fine-grained Alignments between Molecules and Texts}
\author{Jiatong Li, Yunqing Liu, Wei Liu, Jingdi Lei, Di Zhang, Wenqi Fan, Dongzhan Zhou, Yuqiang Li, and Qing Li

\IEEEcompsocitemizethanks{
\IEEEcompsocthanksitem J. Li, Y. Liu, W. Fan, and Q. Li are with the Department of Computing, The Hong Kong Polytechnic University. E-mail:  \{jiatong.li, yunqing617.liu\}@connect.polyu.hk, wenqifan03@gmail.com, csqli@comp.polyu.edu.hk. 
\IEEEcompsocthanksitem W. Liu is with Shanghai Jiao Tong University. E-mail: captain.130@sjtu.edu.cn.
\IEEEcompsocthanksitem J. Lei, D. Zhang, D. Zhou and Y. Li is with Shanghai AI Lab. E-mail: kyrie.jd.lei@gmail.com, di.zhang@ustc.edu, \{zhoudongzhan, liyuqiang\}@pjlab.org.cn.

}
\thanks{(Corresponding authors: Wenqi Fan, Dongzhan Zhou, Yuqiang Li, and Qing Li.)}
}

\markboth{IEEE TRANSACTIONS ON KNOWLEDGE AND DATA ENGINEERING, SUBMISSION 2025}%
{Shell \MakeLowercase{\textit{et al.}}: Bare Demo of IEEEtran.cls for Computer Society Journals}

\maketitle

\begin{abstract}
Molecule discovery is a pivotal research field, impacting everything from medicine to materials. 
Recently, Large Language Models (LLMs) have been widely adopted in molecular understanding and generation, serving as a bridge between the molecular space and the natural language space, yet the alignment between molecules and their corresponding captions remains a significant challenge.
Previous endeavors typically treat molecules as monolithic inputs, lacking an intermediate reasoning process and sacrificing explainability.
In this work, we define fine-grained alignments as the precise correspondence between a molecule’s sub-structures and the textual phrases that explain their properties. These alignments are crucial for LLMs to understand molecules in a more accurate and explainable manner.
Normally, such fine-grained alignments require expert annotation, which is both costly and time-consuming.
To allow LLMs to automatically label and learn the fine-grained alignments, we propose MolReFlect, a novel teacher-student framework, where a teacher LLM first generates and refines mappings between caption phrases and SMILES substructures and then explicitly teaches these detailed alignments to a student LLM.
Experimental results demonstrate that MolReFlect enables LLMs to significantly outperform previous baselines, achieving the state-of-the-art performance in the molecule-caption translation task. 
Our codes are available via: \url{https://github.com/phenixace/MolReFlect}.
\end{abstract}


\section{Introduction}
\label{Introduction}
Molecules are the fundamental units of matter, which normally consist of atoms held together by chemical bonds.
In various chemical and biological processes, molecules play a critical role in participating in reactions \cite{grozinger2002deacetylase}, transmitting signals \cite{raymo2001signal}, and maintaining the structure and function of living organisms \cite{konieczny2023structure}.
It is important to study molecules and their properties, which could benefit a wide range of fields, including Pharmacology \cite{keiser2010chemical}, Agriculture \cite{twyman2003molecular, basaran2008plant}, Material science \cite{higuchi2023material}, and Environmental Ecology \cite{nguyen2017aptamer, valavanidis2006molecular}.

As molecules can be represented by textual systems like SMILES \cite{weininger1988smiles} and SELFIES \cite{krenn2020self}, it is natural to adopt Large Language Models (LLMs) in molecule-related tasks \cite{zhang2024chemllm}. Specifically, LLMs could predict the molecular properties based on the SMILES or SELFIES representations and generate molecules with desired properties, making them helpful assistants for chemists. Correspondingly, Edwards et al. \cite{edwards-etal-2022-translation} propose the molecule-caption translation task to bridge the gap between molecular and natural language space, which contains two sub-tasks, molecule captioning (Mol2Cap) and text-based de novo molecule generation (Cap2Mol).
Successively, MolReGPT \cite{li2023empowering} and ICMA \cite{li2024large} introduced in-context learning to help LLMs understand molecules, while MoMu \cite{su2022molecular} and MolCA \cite{liu2023molca} introduced extra molecular modalities.
However, challenges still exist in the alignments between molecules and texts.

\begin{figure*}
    \centering
    \vskip -0.15in
    \includegraphics[width=0.8\linewidth]{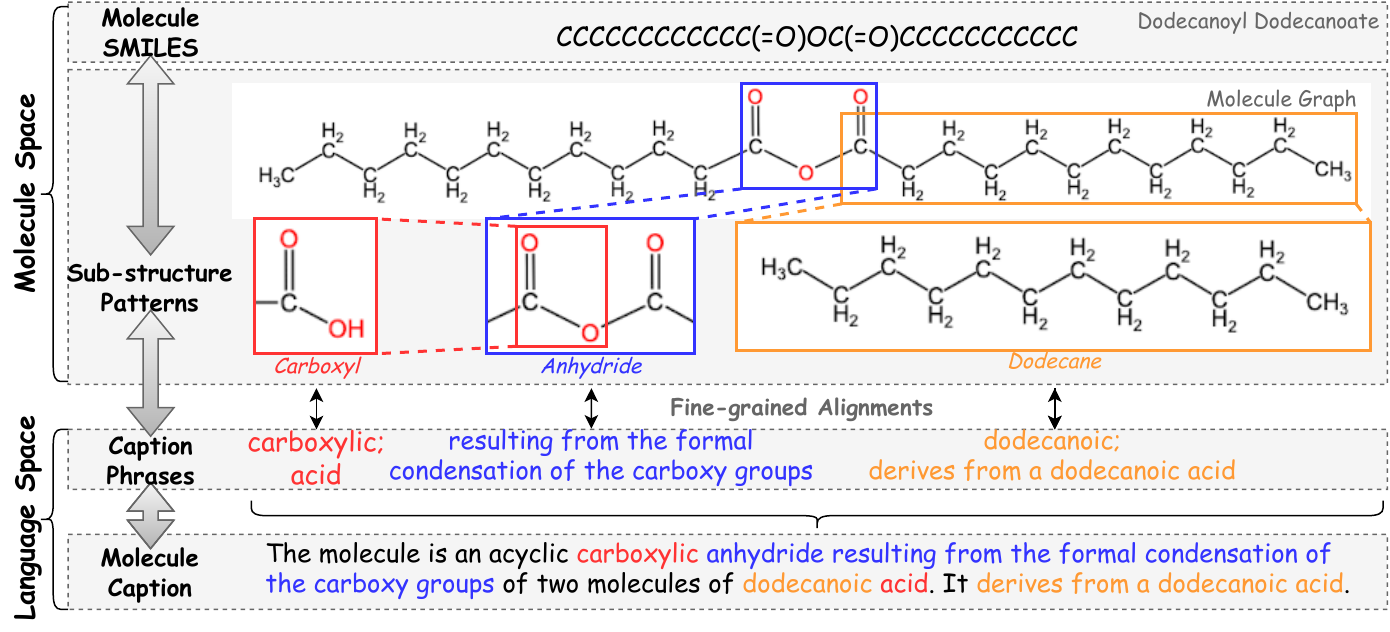}
    \vskip -0.1in
    \caption{An illustration of fine-grained alignments. The sub-structures and their corresponding caption phrases are colored with the same colors to signify the correspondence. Here, the molecule \textit{Dodecanoyl Dodecanoate} (\texttt{CCCCCCCCCCCC(=O)OC(=O)CCCCCCCCCCC})
    is the reaction product of two dodecanoic acids. Thus, it has an anhydride group, and there are 12 carbon atoms on each side of the central oxygen atom.}
    \label{fig:intro}
    \vskip -0.1in
\end{figure*}

Current methods still treat the molecule as a monolithic textual string or molecular graph, neglecting the granularity of alignments and the explainability of their methods. Specifically, sub-structures in the molecule, such as functional groups, can determine the characteristics of the molecule described in the molecule caption. Similarly, characteristics described in the molecule caption can also refer to specific sub-structures of the molecule. 
For example, as shown in Figure \ref{fig:intro}, the molecule \textit{Dodecanoyl Dodecanoate} is the reaction production of the formal condensation of two dodecanoic acids, which turns two carboxyls (\texttt{RC(=O)OH}) into an anhydride (\texttt{RC(=O)OC(=O)R}). Therefore, it has an anhydride group and there are 12 carbon atoms on each side of the central oxygen atom. If LLMs could recognize these patterns, they are more likely to make accurate predictions.
In this case, we introduce fine-grained alignment as the precise correspondence between a molecule’s sub-structures and the textual phrases that elucidate their properties, thereby providing richer information for LLMs to understand molecules.
Nevertheless, few works have paid attention to refining the fine-grained alignments, as they often require domain experts for the annotation, which is both costly and time-consuming.

To resolve the above challenges, two key components are needed: 1) automatic, high-quality annotations of fine-grained alignments, and 2) a robust method to integrate fine-grained alignments into LLMs.
Therefore, we propose \textbf{MolReFlect}, a novel teacher-student framework inspired by reflection tuning \cite{li2024selective}, enabling a larger teacher LLM to collaborate with smaller student LLMs for the molecule-caption translation task with in-context fine-grained alignments. 
MolReFlect includes three stages: \textbf{Zero-shot Alignment Extraction}, \textbf{In-Context Selective Reflection}, and \textbf{Chain-of-Thought In-Context Molecule Tuning (CoT-ICMT)}.
Initially, a larger teacher LLM generates zero-shot alignments by extracting important phrases from SMILES representations or molecule captions and maps them to corresponding characteristics or sub-structure patterns in a zero-shot manner.
To improve the quality of the alignments, we further introduce In-Context Selective Reflection, which first retrieves similar samples and their corresponding zero-shot alignments as in-context few-shot examples so that the teacher LLM can reflect on them and then refine its responses. 
Following this, a student LLM selects between the zero-shot alignments and reflected alignments with lower perplexities to ensure that they could understand the knowledge taught by the teacher LLM and further relieve the noises.
Finally, to help the student LLMs better reason and learn from fine-grained alignments, we develop a new fine-tuning paradigm, Chain-of-Thought In-Context Molecule Tuning (CoT-ICMT) by reformatting the context examples within a thought chain of $input \rightarrow alignments \rightarrow target$.

To verify the effectiveness of our method and study the mechanisms behind MolReFlect, we design a series of experiments on a wide range of datasets, including ChEBI-20 \cite{edwards-etal-2022-translation}, PubChem \cite{liu2023molca}, and MoleculeNet \cite{wu2018moleculenet}.
Experimental results have shown that our method achieves the state-of-the-art (SOTA) performance against all the baselines in both the Mol2Cap and Cap2Mol tasks. 
Meanwhile, the ablation study and extensive analysis also help demonstrate the effectiveness and mechanism of MolReFlect components. 
Furthermore, detailed case studies and error analysis clearly explain how fine-grained alignments work to improve the overall performance in the molecule-caption translation task.
To summarize, our contributions mainly lie in:
\begin{itemize}[itemsep=2pt,topsep=0pt,parsep=0pt]
    \item MolReFlect is the first to explore the fine-grained alignments between molecules and texts, achieving the SOTA performance in the molecule-caption translation task.
    \item By integrating transparent, fine-grained alignments as reasoning intermediates, MolReFlect contributes to a more explainable framework, helping LLMs better understand the translation process between molecules and texts.
    \item MolReFlect can work with general LLMs without domain-specific pre-training, providing a viable solution to relieve the data hunger in the biochemical field.
\end{itemize}

\section{Related Work}
\label{sec:relatedwork}

LLMs have demonstrated great potential in Molecule Discovery, including molecule understanding~\cite{qian2023can}, optimization~\cite{ye2023drugassist}, and generation~\cite{irwin2022chemformer}. 
To align molecule representation with natural language texts,
the MolT5 study first proposed the molecule-caption translation task, introducing a new dataset, ChEBI-20, with pairs of molecule SMILES representations and their textual captions that describe the structural patterns and chemical properties~\cite{edwards2021text2mol}. 
Subsequent research has intensified the focus on this task, branching out in two primary directions.

On one trajectory, the research leverages the in-context learning capability of LLMs and the similarity principle of molecules to help LLMs learn the molecule-text alignment in context~\cite{li2023empowering}. Advancing this approach, ICMA has developed In-Context Molecule Tuning (ICMT), significantly enhancing the capabilities of LLMs in the molecule-caption translation task and reducing the reliance on domain-specific pre-training~\cite{li2024large}.
Concurrently, the other works involve incorporating additional information from different modalities into LLMs.
For instance, MoMu~\cite{su2022molecular} adopts contrastive learning to align the output distribution of the text encoder with the graph encoder, 
while the CLIP structure~\cite{radford2021learning} is not good at generative tasks.
In this case, MolCA~\cite{liu2023molca} introduce the 2D molecular graph with a Q-Former structure~\cite{li2023blip} to enhance the performance of LLMs in the molecule captioning task.
Meanwhile, 3D-MoLM~\cite{li2024towards} adopts the similar Q-Former structure, introducing 3D molecule information to LLMs. 
However, these methods still treat molecules as monolithic inputs, suffering from the necessity for additional modality alignment steps, low precision, and poor interpretability. 

\newtheorem{assumption}{Assumption}
\newtheorem{theorem}{Theorem}
\newtheorem{lemma}{Lemma}
\newtheorem{remark}{Remark}

\newtcbtheorem[number within=section]{boxedtheorem}{Theorem}%
{enhanced,
  colback=white,
  colframe=black,
  coltitle=black,
  fonttitle=\bfseries,
  sharp corners,
  boxrule=0.8pt,
  attach boxed title to top left={yshift=-2mm,xshift=3mm},
  boxed title style={colback=white,boxrule=0pt},
  before skip=10pt,after skip=10pt
}{th}

\section{Preliminaries}
\begin{figure*}
    \centering
    \includegraphics[width=0.75\linewidth]{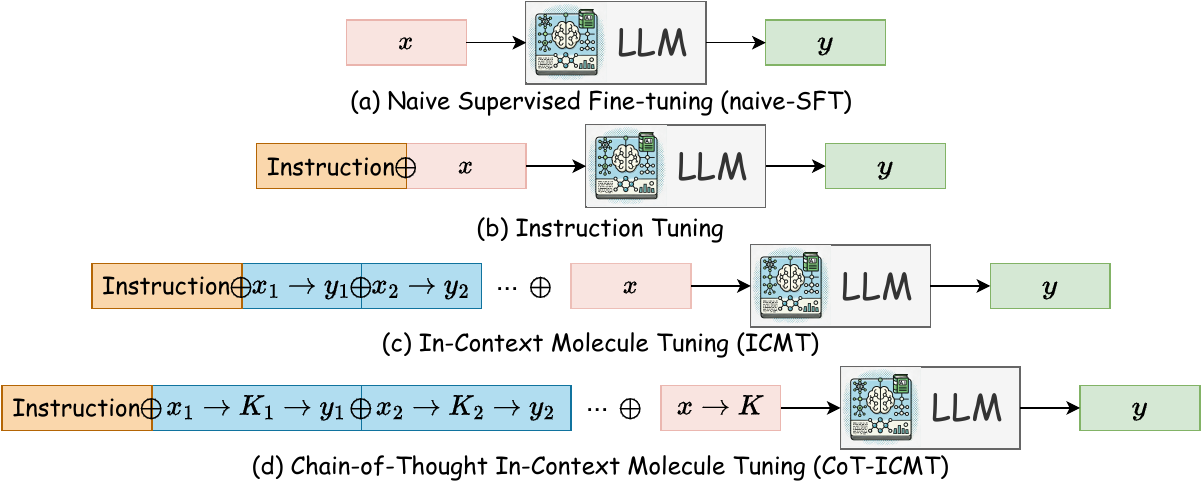}

    \caption{Comparison of fine-tuning paradigms, including (a) Naive Supervised Fine-tuning (naive-SFT), (b) Instruction Tuning \cite{wei2021finetuned}, (c) In-Context Molecule Tuning (ICMT) \cite{li2024large}, and (d) Chain-of-Thought In-Context Molecule Tuning (CoT-ICMT).}
    \label{fig:loss}

\end{figure*}
We demonstrate the differences among three previous fine-tuning paradigms illustrated in Figure \ref{fig:loss} (a-c), including Naive-Supervised Fine-tuning, Instruction Tuning \cite{wei2021finetuned}, and In-context Molecule Tuning \cite{li2024large}.
Generally, given an LLM and its parameters $\theta$, supposing that the training set is $D$ and $(x,y) \in D$ denotes a molecule-caption pair from the training set, the LLM ought to generate the response $y \sim p_\theta(.|x)$ based on the input text $x$. Notably, in this paper, $x$ refers to both the input molecule and input caption, while $y$ refers to the corresponding target caption and target molecule. Naive supervised fine-tuning (naive-SFT) learns the mapping from input to target $x \rightarrow y$ directly. Accordingly, the loss function of naive-SFT could be represented as follows:
\begin{align}
    L^{nft}(\theta) =  \underset{(x,y)\in D}{\sum}\left[-\log p_\theta(y|x)\right],
\end{align}

Different from naive-SFT, Instruction Tuning \cite{wei2021finetuned} introduces instructions to guide the generation of LLMs. Normally, instructions contain task-related information such as role identification and additional knowledge. Formally, given the task instruction $I$, the loss function of Instruction Tuning can be denoted as:
\begin{align}
    L^{it}(\theta) =  \underset{(x,y)\in D}{\sum}\left[-\log p_\theta(y|x, I)\right],
\end{align}
Inspired by In-Context Tuning \cite{chen2022meta}, 
Li et. al. \cite{li2024large} take a step further and propose In-Context Molecule Tuning (ICMT), which introduces n similar molecule-caption examples $\{(x_i,y_i)\}_{i=1}^n \subset D$. Therefore, the LLM will make predictions based on the text content $C_{x\rightarrow y}$ $=\{\mathcal{P}(x_i,y_i)\}_{i=1}^n$ and the mappings $F_{x\rightarrow y} = \{f_i:= x_i \rightarrow y_i\}_{i=1}^n$ behind the context examples, where $\mathcal{P}$ denotes the prompt template. Thus, as illustrated in Figure \ref{fig:loss} (c), the loss function of ICMT can be written as:
{
\begin{align}\small
    L^{icmt}(\theta)=\underset{(x,y)\in D}{\sum}\left[-\log p_\theta(y|x, [C_{x\rightarrow y}, F_{x\rightarrow y}], I)\right].
    \label{eq3}
\end{align}
}

\section{MolReFlect}
\label{sec:methodlogy}
MolReFlect employs a teacher-student architecture, where an advanced (larger and more powerful) language model serves as the teacher, and less sophisticated (smaller but more efficient) language models act as the students.
The rationale for this architecture is twofold:

First, annotating high-quality fine-grained alignments requires a model with superior reasoning capabilities and a broader knowledge base, which the larger teacher model can provide.
Second, although the teacher may still produce errors, the student models are designed to critically evaluate and integrate the teacher’s outputs, enabling them to learn selectively from correct alignments and develop robustness against the teacher’s potential mistakes.

\begin{figure*}
    \centering
    \vskip -0.1in
    \includegraphics[width=0.95\linewidth]{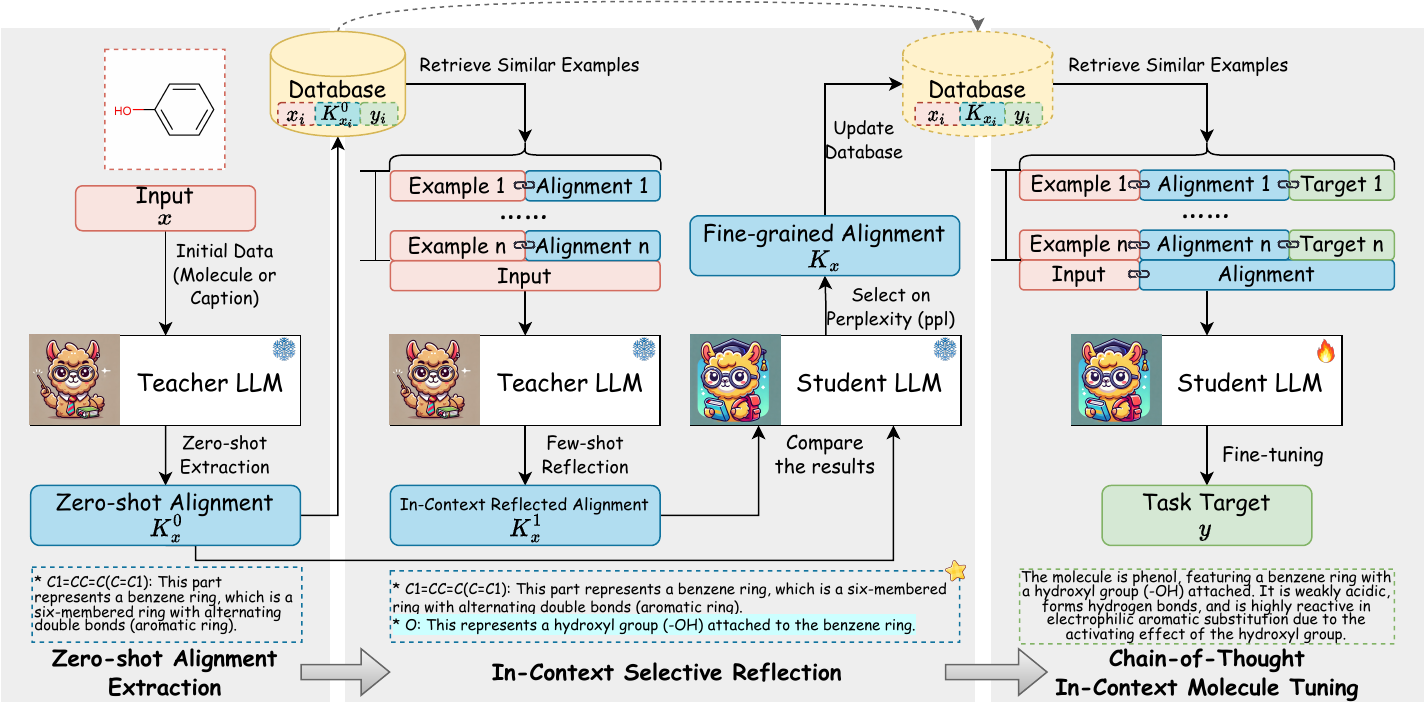}
    \vskip -0.1in
    \caption{The overall framework of MolReFlect consists of three stages, Zero-shot Alignment Extraction, In-Context Selective Reflection, and Chain-of-Thought In-Context Molecule Tuning.}
    \label{fig:model}

\end{figure*}

Therefore, as depicted in Figure \ref{fig:model}, MolReFlect is organized into three principal stages: Zero-Shot Alignment Extraction, In-Context Selective Reflection, and CoT-ICMT, where the teacher model leverages its superior capabilities to generate initial alignments, and the student model critically evaluates and learns from them.

\subsection{Zero-shot Alignment Extraction} 
Instead of directly learning the mappings from molecules to captions, MolReFlect aims to extract fine-grained alignments $K$ between the molecule SMILES strings and captions, 
thereby learning the mapping chains $x \rightarrow K \rightarrow y$.
The fine-grained alignments should indicate the correspondence between specific sub-structures in the molecule (e.g., functional groups, atoms) and the phrases or words in the caption.
To leverage the inherent chemical knowledge of LLMs while avoiding the prohibitive cost of creating a labeled alignment dataset, we developed a zero-shot prompting strategy for the teacher LLM to conduct extraction via chain-of-thought (CoT) reasoning \cite{wei2022chain}. This allows the teacher LLM to distill critical sub-structures and phrases from the molecule SMILES representations and captions, respectively, offering implications for final predictions.
Formally, we have:
\begin{align}
    K^0 = p_{\theta_T}(.|x, I),
\end{align}
\noindent where $\theta_T$ represents the parameters of the larger teacher LLM, $I$ is the CoT instruction, and $K^0$ signifies the textual alignments extracted in a zero-shot manner from the molecule caption and SMILES string.

\subsection{In-Context Selective Reflection}
Despite the powerful reasoning capabilities of LLMs, they can still generate answers with hallucinations~\cite{yao2023llm}, which can introduce noise into the zero-shot alignments.
To mitigate these potential noise and enhance the quality of zero-shot alignments,
we propose a strategy that allows the teacher LLM to self-reflect on the zero-shot extraction results through in-context few-shot learning, where the previous zero-shot alignments are retrieved by similarity and serve as context examples for reflection. 
This context examples allow the teacher LLM to calibrate its understanding of molecules, effectively filtering out noise and improving the fidelity of the final output.

For the retrieval process, we follow the retrieval strategy adopted in \cite{li2024large}. Given the input $x$, the context examples can be formed as follow:
\begin{align}
    C_x = \{(x_1,K^0_1),(x_2,K^0_2), ..., (x_n,K^0_n)\},
\end{align}
In this case, we can obtain the in-context reflected alignments $K^1$ through the teacher LLM:
\begin{align}
    K_1 = p_{\theta_T}(.|x, C_x, I), 
\end{align}
Notably, the zero-shot alignments of the current input are not wrapped into the context to prevent the LLM from directly repeating it, and maintain consistent prompt formats across all instances.

However, context examples might still contain noise that could misguide the reflection process, potentially leading to a decline in quality of reflected alignments $K^1$ compared to the zero-shot alignments $K^0$. Furthermore, the alignments generated by the teacher LLM can sometimes be too complex for the smaller student LLM to comprehend.
Therefore, it is necessary to choose the better one between $K^0$ and $K^1$.
To avoid possible information leaks, an unsupervised metric is required for selection. Specifically, we adopt the perplexity $\rm{ppl}$ as the metric from the information theory perspective:
\begin{align}
    {\rm ppl}(K_x,x) = \log\left[-p_{\theta_S}(K_x|x)\right],
\end{align}
\noindent where $x$ is the input, $K_x$ denotes the corresponding alignments, and $\theta_S$ is the original parameters of the smaller student LLM. Normally, low-perplexity sequences tend to exhibit higher semantic validity and predictive utility \cite{li2024selective,wangself}. Therefore, we could make the following assumption:
\begin{assumption}[perplexity / usefulness correlation]
\label{asm:usefulness}
Let $\theta_S$ denote the student model. For any input $x$ and any candidate alignment $K$ in the candidate set $\mathcal{K}(x)$, the student-assigned probability $p_{\theta_S}(K\mid x)$ is positively correlated with the ``usefulness'' of $K$ for predicting $Y$ from $(X,K)$. Concretely, we assume there exists a (nonincreasing) function $\Phi:\ (0,1]\to\mathbb{R}$ such that for all $x$ and $K$,
\[
\mathbb{E}_{Y\mid x}\big[-\log p^*(Y\mid x,K)\big]
\;=\;\Phi\big(p_{\theta_S}(K\mid x)\big),
\]
with $\Phi$ monotone nonincreasing. Here $p^*$ denotes the true data-generating conditional distribution.
\end{assumption}
Based on this assumption, Theorem IV.1 proves that the selection based on perplexity can reduce the successive training loss of student LLMs. 

Notably, we adopt Galactica-125M \cite{taylor2022galactica} as the student LLM for perplexity calculation due to its chemical expertise and manageable size, which allows for efficient metric computation.
Between the zero-shot alignment and the in-context reflected alignment, the one with lower perplexity will be selected as the final fine-grained alignments:
\begin{align}
    K &= \left\{
    \begin{array}{ll}K^0 &\ \text{if ${\rm ppl}(K^0,x) < {\rm ppl}(K^1,x)$}, \\ K^1 &\ \text{elsewise}, \end{array}\right.
\end{align}

\begin{figure*}[t]
\centering
\begin{minipage}{0.9\textwidth}
\begin{boxedtheorem}{Perplexity-based selection reduces expected training cross-entropy}{ppl_selection}
Let the data distribution be $(X,Y)\sim\mathcal{D}$. For each $x$ let $\mathcal{K}(x)$ be a finite candidate set of alignments. Define the per-sample choice
\[
K^\star(x)\;=\;\arg\max_{K\in\mathcal{K}(x)} p_{\theta_S}(K\mid x),
\]
(i.e. equivalently $\arg\min$ perplexity $-\log p_{\theta_S}(K\mid x)$). Under Assumption \ref{asm:usefulness}, for any (learnable) model family $p_\theta(y\mid x,K)$ and any fixed $\theta$,
\[
\begin{aligned}
\mathbb{E}_{(X,Y)\sim\mathcal{D}}\!\big[-\log p_\theta(Y\mid X,K^\star(X))\big]
\;\le\; 
\mathbb{E}_{(X,Y)\sim\mathcal{D}}\!\big[-\log p_\theta(Y\mid X,K)\big],\
\forall\, K\in\mathcal{K}(\cdot).
\end{aligned}
\]
i.e. the per-sample choice $K^\star$ does not increase (and under strict monotonicity strictly decreases) the expected negative log-likelihood compared to using any fixed candidate alignment.

\begin{proof}
Fix $x$. For any $K\in\mathcal{K}(x)$, define the conditional expected loss
\[
\mathcal{L}(x,K;\theta)\;=\;\mathbb{E}_{Y\mid x}\big[-\log p_\theta(Y\mid x,K)\big].
\]
By Assumption \ref{asm:usefulness} we have $\mathcal{L}(x,K;\theta)$ is a (weakly) decreasing function of $p_{\theta_S}(K\mid x)$ when $\theta$ is near the true model family or when $p_\theta$ tracks $p^*$ in the sense used to formulate the assumption. Consequently, the $K$ that maximizes $p_{\theta_S}(K\mid x)$ also (pointwise in $x$) minimizes $\mathcal{L}(x,K;\theta)$ over $\mathcal{K}(x)$. Thus for every $x$,
\[
\mathcal{L}(x,K^\star(x);\theta)\;\le\;\mathcal{L}(x,K;\theta),\qquad\forall K\in\mathcal{K}(x).
\]
Taking expectation over $X\sim\mathcal{D}_X$ yields the stated inequality.
\end{proof}
\end{boxedtheorem}
\end{minipage}
\end{figure*}

\subsection{Chain-of-Thought In-Context Molecule Tuning}
\begin{figure*}[t]
\centering
\begin{minipage}{0.9\textwidth}
\begin{boxedtheorem}{Fine-grained alignments reduce information-theoretic variance}{cot}
Let $(X,K,Y)$ be jointly distributed, where $K$ denotes a CoT-style intermediate (fine-grained alignments). Then
\[
H(Y\mid X,K)\le H(Y\mid X),
\]
where $H(\cdot)$ denotes Shannon entropy. Consequently, the Bayes-optimal expected negative log-likelihood obtainable by a model that can condition on $(X,K)$ is at most the Bayes-optimal expected negative log-likelihood obtainable when conditioning only on $X$.

\begin{proof}
The identity
\[
H(Y\mid X) \;=\; H(Y\mid X,K) + I(Y;K\mid X)
\]
is standard (chain rule for entropy), and mutual information $I(Y;K\mid X)\ge 0$. Rearranging yields $H(Y\mid X,K)\le H(Y\mid X)$. Since the minimum achievable expected negative log-likelihood (cross-entropy) equals the conditional entropy of $Y$ given the conditioning variables when the model class is sufficiently rich to represent the true conditional distribution, the inequality implies the stated bound on achievable Bayes risk.
\end{proof}
\end{boxedtheorem}
\end{minipage}
\end{figure*}

While leveraging fine-grained alignments as context for a large teacher LLM to generate predictions directly is technically feasible, this approach has significant limitations. Because the teacher LLM is not pre-trained on domain-specific chemical corpora, it lacks familiarity with the desired output distribution of molecular datasets, leading to generations that often fail in structure or semantics. 
Meanwhile, fine-tuning such a large teacher LLM is also computationally expensive and inefficient.

To address these limitations, we fine-tune a smaller student LLM to learn from the teacher’s fine-grained alignments. In this setup, our primary objective is to leverage the student model’s reasoning capabilities, not its chemical knowledge. This focus on reasoning, rather than domain expertise, leads us to select models in the 7-8B parameter range, which offer an optimal balance between capability and training efficiency.

In contrast to In-Context Molecule Tuning \cite{li2024large}, CoT-ICMT organizes the fine-grained alignments of both the input $x$ and the context examples $C_x$ into the CoT format. Theorem IV.2 proves that this CoT format can reduce information theoretic variance with the fine-grained alignments, thereby facilitating better performance. During the process of CoT-ICMT, top-$n$ similar examples are retrieved via the same retrieval strategies in \cite{li2024large} and then organized into the context with the CoT format to fine-tune the parameters of the smaller student LLM.
Formally, similar to Eq.~\ref{eq3}, the loss function can be represented as follows:
{
\begin{align}\small
    L&^{cot-icmt}(\theta) = \nonumber\\
    &\underset{(x,y)\in D}{\sum}\left[-\log p_\theta(\!y|x,\!K_x\!,\![C_{x\!\rightarrow\! K_x\!\rightarrow\!y}\!,\! F_{x\!\rightarrow\!K_x\rightarrow\!y}],\!I\!)\right],
\end{align}
}
where $K_x$ denotes the fine-grained alignments of input $x$, $C_{x\rightarrow K_x\rightarrow y}$ $=\{\mathcal{P}(x_i, K_{x_i}, y_i)\}_{i=1}^n$ represents the text content of context examples organized by the CoT format prompt $\mathcal{P}$, and $F_{x\rightarrow K_x\rightarrow y}=\{f_i:=x_i \rightarrow K_{x_i} \rightarrow y_i\}_{i=1}^n$ denotes the mapping chains underlying the context examples, which map the original inputs to the fine-grained alignments and then further map the fine-grained alignments to the final targets.

\section{Experiments}
\label{sec:Experiments}
In this section, we present our experiment setups and compare MolReFlect against existing baselines. 
Then, we conduct a series of ablation experiments to validate our proposed framework, focusing on the following specific research questions: \textbf{(RQ1)} \uline{Do fine-grained alignments improve the performance in the molecule-caption translation task, and if so, how?} \textbf{(RQ2)} \uline{Why is it necessary to reflect and select between the zero-shot alignments and in-context reflected alignments?} \textbf{(RQ3)} \uline{What is the necessity of adopting a teacher-student framework?}

\begin{table*}[htbp]
    \centering
    \caption{Overall performance comparison for the Mol2Cap task on the ChEBI-20 dataset (\textbf{Best}, \underline{Second Best}). Except for MolReGPT, all the other methods involve fine-tuning LLMs on the ChEBI-20 dataset. }
    \resizebox{1.35\columnwidth}{!}{
    \begin{tabular}{c|c|c|c|c|c|c}
    \toprule
    Method & BLEU-2$\uparrow$ & BLEU-4$\uparrow$ & ROUGE-1$\uparrow$ & ROUGE-2$\uparrow$ & ROUGE-L$\uparrow$ & METEOR$\uparrow$ \\
    \midrule
    MolT5-large & 0.594 & 0.508 & 0.654 & 0.510 & 0.594 & 0.614 \\
    MolReGPT & 0.607 & 0.525 & 0.634 & 0.476 & 0.562 & 0.610\\
    MolCA & 0.639 & 0.555 & \underline{0.697} & 0.558 & \underline{0.636} & \underline{0.669} \\
    BioT5 & 0.635 & 0.556 & 0.692 & \underline{0.559} & 0.633 & 0.656 \\
    ICMA& \underline{0.651}	& \underline{0.581}	& 0.686 &	0.550 & 0.625 & 0.661 \\  
    \midrule
    \textbf{MolReFlect} &  \textbf{0.676} & \textbf{0.608}&\textbf{0.703} & \textbf{0.571} & 
    \textbf{0.644} & \textbf{0.680}\\
    \bottomrule
    \end{tabular}
    }
    \label{tab:m2c_base}
\end{table*}

\begin{table*}[htbp]
    \centering
    \caption{Overall performance comparison for the Cap2Mol task on the ChEBI-20 dataset (\textbf{Best}, \underline{Second Best}). Except for MolReGPT, all the other methods involve fine-tuning LLMs on the ChEBI-20 dataset.}
    \resizebox{1.5\columnwidth}{!}{
    \begin{tabular}{c|c|c|c|c|c|c|c}
    \toprule
    Method & BLEU$\uparrow$ & EM$\uparrow$ & Levenshtein$\downarrow$ & MACCS FTS$\uparrow$ & RDK FTS$\uparrow$ & Morgan FTS$\uparrow$ & Validity$\uparrow$ \\
    \midrule
    MolT5-large & 0.854 & 0.311 & 16.07 & 0.834 & 0.746 & 0.684 & 0.905\\
    MolReGPT & 0.857 & 0.280 & 17.14 & 0.903 & 0.805 & 0.739 & 0.899\\
    BioT5 & \underline{0.867} & 0.413	& \underline{15.10} &	0.886 &	0.801 &	0.734 & \textbf{1.000}   \\
    ICMA& 0.855 & \underline{0.460} & 18.73 & \underline{0.916} & \underline{0.837} & \underline{0.789} & 0.958  \\
    \midrule
    \textbf{MolReFlect}& \textbf{0.903}& \textbf{0.510} &	\textbf{11.84}	&\textbf{0.929}	& \textbf{0.860}	& \textbf{0.813}	&\underline{0.977}\\
    \bottomrule
    \end{tabular}
    }
    \label{tab:c2m_base}
    \vskip -0.1in
\end{table*}

\subsection{Experiment Setups} 

\noindent\textbf{Implementation Details.} For the larger teacher LLM, we adopt the powerful Llama-3-70B-Instruct model \cite{dubey2024llama}, as its competitive performance against GPT-4 \cite{achiam2023gpt} makes it well-suited for the role of teacher.
For the smaller student LLM, we mainly adopt Mistral-7B-Instruct-v0.2 (Mistral-7B for short) \cite{jiang2023mistral} for fair comparisons to ICMA \cite{li2024large}.
In this work, we focus on the ChEBI-20 dataset \cite{edwards-etal-2022-translation} and all the experiments are conducted on Nvidia RTX A6000 and A100 GPUs.
We adopt the vllm\footnote{\url{https://github.com/vllm-project/vllm}} framework to deploy the int4 quantized llama-3-70B-Instruct on the local devices as OpenAI compatible server\footnote{\url{https://platform.openai.com/docs/guides/chat-completions}}. At the same time, we utilize the huggingface transformers\footnote{\url{https://huggingface.co}} and Lora adapters \cite{hu2021lora} for the fine-tuning process.

\begin{table}[htbp]
    \centering
    \caption{Hyper-parameters for the larger teacher LLM.}
    \resizebox{0.5\columnwidth}{!}{
    \begin{tabular}{c|c}
    \toprule
    Item & Value \\
    \midrule
    int4 & True \\
    temperature & 0.75 \\
    top\_p & 0.85 \\
    top\_k & 40 \\
    num\_return\_sequences & 1 \\
    max\_new\_tokens & 512 \\
    number-of-examples & 2 \\
    \bottomrule
    \end{tabular}
    }
    \label{tab:Hyper1}
\end{table}

\begin{table}[htbp]
    \centering
    \caption{Hyper-parameters for the smaller student LLM.}
    \resizebox{0.5\columnwidth}{!}{
    \begin{tabular}{c|c}
    \toprule
    Item & Value \\
    \midrule
    macro batch size & 32  \\
    micro batch size & 1  \\
    steps & 8000 \\
    warm-up steps & 1000 \\
    cutoff length & 4096 \\
    number-of-examples & 2 \\
    learning rate & 2e-4 \\
    \midrule
    lora\_r & 32\\
    lora\_alpha & 64 \\
    lora\_dropout & 0.1 \\
    int8 & True \\
    fp16 & True \\
    \midrule
    temperature & 0.75 \\
    top\_p & 0.85 \\
    top\_k & 40 \\
    num\_return\_sequences & 1 \\
    max\_new\_tokens & 512 \\
    \bottomrule
    \end{tabular}
    }
    \label{tab:Hyper2}
\end{table}

\noindent\textbf{Hyper-parameters.} For reproduction, we list all the hyper-parameters used in our framework, including Table \ref{tab:Hyper1} for the prompting of the teacher LLM and Table \ref{tab:Hyper2} for the fine-tuning and testing of the student LLM. Notably, we incorporate $n=2$ examples in both in-context selective reflection and Chain-of-Thought In-Context Molecule Tuning. Furthermore, the Llama-3-70B-Instruct is int4 quantized to allow inference on a single NVIDIA A6000 GPU for data-parallel acceleration, while the Mistral-7B-Instruct-v0.2 is int8 and fp16 quantized during the fine-tuning process. We keep similar generation parameters for both the teacher LLM and the student LLM.

\noindent\textbf{Metrics.} Regarding the evaluation metrics, we adopt the same settings as ICMA. We employ translation metrics for the Mol2Cap task, including BLEU-2,4 scores, ROUGE-1,2,L scores, and METEOR scores. Higher values in these metrics indicate that the generated molecule captions are more aligned with the ground truth.
For the Cap2Mol task, we employ a combination of translation and molecule-specific metrics for evaluation, which includes BLEU, Exact Match, Levenshtein, three Molecule Fingerprints scores, and a validity score. Except for the Levenshtein score, where a lower value is preferable, higher scores across these metrics generally signify better model performance.

\subsection{Overall Performance Comparison} 
We compare our method with baseline models across the two sub-tasks of the ChEBI-20 dataset. Specifically, we select MolT5-large~\cite{edwards-etal-2022-translation}, MolReGPT~\cite{li2023empowering}, MolCA (for the Mol2Cap task only)~\cite{liu2023molca}, BioT5~\cite{pei2023biot5}, and ICMA~\cite{li2024large} as the baseline models. 

\noindent\textbf{Mol2Cap:} As indicated in Table \ref{tab:m2c_base}, MolReFlect achieves the best performance across all evaluation metrics. Significantly, MolReFlect obtains a BLEU-2 score of 0.676 and a BLEU-4 score of 0.608, representing improvements of 3.8\% and 4.6\% over ICMA, while maintaining superior ROUGE scores.
In comparison to domain-specific pre-training approaches such as BioT5 and multi-modal strategies like MolCA, MolReFlect still exhibits superior performance using a general-purpose LLM without any extra domain pre-training or modality alignment stages, thereby underscoring the effectiveness of our framework.

\noindent\textbf{Cap2Mol:} As evidenced in Table \ref{tab:c2m_base}, MolReFlect exhibits superior performance on the Cap2Mol task. Compared to previous baselines such as ICMA, MolReFlect achieves a BLEU score of 0.903 and generates a remarkable 51\% Exact Matched molecules while obtaining a lower Levenshtein score. Moreover, MolReFlect also achieves the highest molecule fingerprint scores, indicating that the generations are more similar to the ground truths. 
As MolReFlect employs SMILES to represent molecules, the validity of generations can be slightly below the 100\% validity of BioT5, which adopts SELFIES as its input. However, using SMILES strings does not require an extension of the tokenizer vocabulary, which preserves the information from pre-training, and
this limitation can be addressed through various sampling and filtering strategies. 

Generally, in both the Mol2Cap and Cap2Mol tasks, MolReFlect consistently demonstrates the SOTA or comparable performance over existing baselines.

\begin{table*}[htbp]
    \centering
    \caption{Ablation analysis of MolReFlect for the Mol2Cap task performance (Mistral-7B-Instruct-v0.2 as backbone). Above: Mistral-7B(naive-SFT) and MolReFlect; Middle: Ablating Context Examples and Fine-grained Alignments; Below: Ablating In-Context Reflection and Selection.}
    \resizebox{1.45\columnwidth}{!}{
    \begin{tabular}{c|c|c|c|c|c|c}
    \toprule
    Method & BLEU-2$\uparrow$ & BLEU-4$\uparrow$ & ROUGE-1$\uparrow$ & ROUGE-2$\uparrow$ & ROUGE-L$\uparrow$ & METEOR$\uparrow$ \\
    \midrule
    Mistral-7B(naive-SFT) & 0.566 & 0.478 & 0.614 & 0.449 & 0.547 & 0.572 \\
    \textbf{MolReFlect}& \textbf{0.676} & \textbf{0.608} & \textbf{0.703} & \textbf{0.571} & \textbf{0.644} & \textbf{0.680}\\
    \midrule
    w/o Examples& 0.617 & 0.539  & 0.657 & 0.510 & 0.593 & 0.623\\
    w/o Alignments & 0.651	& 0.581	& 0.686 &	0.550 & 0.625 & 0.661 \\
    \midrule
    w/o In-Context Reflection & 0.648	&0.580	&0.700(8)	&0.568(3)	&0.640(7)	&0.678\\
    w/o Selection & 0.672 & 0.604 &0.701(1)& 0.568(1)	&0.640(9)&	0.677\\
    \bottomrule
    \end{tabular}
    }
    \label{tab:m2c_iter}
\end{table*}

\begin{table*}[htbp]
    \centering
    \caption{Ablation analysis of MolReFlect for the Cap2Mol task performance (Mistral-7B-Instruct-v0.2 as backbone). Above: Mistral-7B(naive-SFT) and MolReFlect; Middle: Ablating Context Examples and Fine-grained Alignments; Below: Ablating In-Context Reflection and Selection.}
    \resizebox{1.55\columnwidth}{!}{
    \begin{tabular}{c|c|c|c|c|c|c|c}
    \toprule
    Method & BLEU$\uparrow$ & EM$\uparrow$ & Levenshtein$\downarrow$ & MACCS FTS$\uparrow$ & RDK FTS$\uparrow$ & Morgan FTS$\uparrow$ & Validity$\uparrow$ \\
    \midrule
    Mistral-7B(naive-SFT) & 0.767 & 0.234	& 27.39 &	0.852 &	0.718 &	0.649 & 0.918   \\
    \textbf{MolReFlect}& \textbf{0.903} & \textbf{0.510} & \textbf{11.84} & \textbf{0.929}	& \textbf{0.860}	& \textbf{0.813}	&0.977 \\
    \midrule
    w/o Examples& 0.886 & 0.430  & 13.99 & 0.916 & 0.828 & 0.775 & 0.981\\
    w/o Alignments & 0.855 & 0.460 & 18.73 & 0.916 & 0.837 & 0.789 & 0.958  \\
    \midrule
    w/o In-Context Reflection &  0.900(3)&	0.502&	11.94	&0.926	&0.855&	0.807&	0.979\\
    w/o Selection & 0.900(1)	&0.496	&12.86	&0.927	&0.858	&0.808	& \textbf{0.980}\\
    
    \bottomrule
    \end{tabular}
    }
    \label{tab:c2m_iter}
\end{table*}

\subsection{Ablation Study \& Discussion}
To enable a better understanding of MolReFlect, we conduct a series of ablation studies to resolve the research questions (\textbf{RQ1-3}) that have been raised for discussion.

\noindent\textbf{RQ1: Do fine-grained alignments improve the performance in the molecule-caption translation task, and if so, how?}

Mathematically, Theorem IV.2 demonstrates that fine-grained alignments reduce the information-theoretic variance during training, thereby improving the stability and accuracy of the final predictions.
Meanwhile, we also conduct an empirical ablation study on MolReFlect by removing context examples or fine-grained alignments, downgrading MolReFlect to Instruction Tuning and ICMA, respectively. Besides, we also provide the naive-SFT performance of Mistral-7B for comparison. 
The corresponding results are presented in Table \ref{tab:m2c_iter} and Table \ref{tab:c2m_iter}.

Overall, the naive-SFT baseline performs the worst because Mistral-7B lacks domain-specific pretraining on chemical corpora.
In contrast, MolReFlect substantially outperforms naive-SFT on both two subtasks, showcasing the effectiveness of our framework.
Meanwhile, when context examples are removed, the performances drop slightly but attain a BLEU-4 score of 0.539 on the Mol2Cap task and a BLEU score of 0.886 on the Cap2Mol task, remaining substantially higher than naive-SFT. 
Notably, in the Cap2Mol task, the Exact Match score nearly doubles compared to naive-SFT, indicating that fine-grained alignments effectively convey molecular structural information to the student LLM. 

Similarly, when fine-grained alignments are removed during the fine-tuning phase, the performance also drops on both Mol2Cap and Cap2Mol tasks. This suggests that context examples can guide student LLMs to learn molecule-text alignments more effectively from fine-grained alignments, leading to better final generations.

Therefore, the experimental results show that the smaller student LLM could learn from fine-grained alignments more effectively via context examples. 

\begin{table*}[htbp]
    \centering
    \vskip -0.15in
    \caption{Performance comparison of prompting strategies for the teacher LLM (Llama-3-70B-Instruct) to perform the Mol2Cap task independently.}
    \vskip -0.1in
    \resizebox{1.5\columnwidth}{!}{
    \begin{tabular}{c|c|c|c|c|c|c|c}
    \toprule
    Method & BLEU-2$\uparrow$ & BLEU-4$\uparrow$ & ROUGE-1$\uparrow$ & ROUGE-2$\uparrow$ & ROUGE-L$\uparrow$ & METEOR$\uparrow$ & AVG IMP \\
    \midrule
    Direct Prompting& 0.071&	0.038	&0.220&	0.093&	0.192&	0.139 & -\\
    Chain-of-Thought&  0.149	&0.075&	0.249	&0.089	&0.204 &0.179 & 41.80\%\\
    \midrule
    Few-shot Prompting& 0.457&	0.389&	0.556	&0.399	&0.492&	0.481 & -\\
    Few-shot Chain-of-Thought& 0.474&	0.382&	0.523&	0.349&	0.449&	0.476 & -4.41\%\\
    \bottomrule
    \end{tabular}
    }
    \label{tab:m2c_pub}
    \vskip -0.1in
\end{table*}

\begin{table*}[htbp]
    \centering
    \caption{Performance comparison of prompting strategies for the teacher LLM (Llama-3-70B-Instruct) to perform the Cap2Mol task independently.}
    \vskip -0.1in
    \resizebox{1.6\columnwidth}{!}{
    \begin{tabular}{c|c|c|c|c|c|c|c|c}
    \toprule
    Method & BLEU$\uparrow$ & EM$\uparrow$ & Levenshtein$\downarrow$ & MACCS FTS$\uparrow$ & RDK FTS$\uparrow$ & Morgan FTS$\uparrow$ & Validity$\uparrow$ & AVG IMP \\
    \midrule
    Direct Prompting & 0.417	& 0.032	& 46.91&	0.711	&0.474	&0.411	&0.666 & -\\
    Chain-of-Thought&  0.380	&0.033	&47.46	&0.708	&0.476	&0.407	&0.683 & -1.05\%\\
    \midrule
    Few-shot Prompting &  0.773	&0.134	&22.53	&0.869	&0.748&	0.679&	0.751 & - \\
    Few-shot Chain-of-Thought&  0.759&	0.129&	23.13&	0.872&	0.752&	0.679&	0.766 & 0.74\%\\
    \bottomrule
    \end{tabular}
    }
    \vskip -0.1in
    \label{tab:c2m_pub}
\end{table*}

\noindent\textbf{RQ2: Why is it necessary to reflect and select between the zero-shot alignments and in-context reflected alignments?}

To answer this question, we ablate MolReFlect by removing the in-context reflection and the selection processes, which is equivalent to replacing the fine-grained alignments with zero-shot alignments and in-context reflected alignments, respectively. The details are shown in the last two rows of Table \ref{tab:m2c_iter} (for the Mol2Cap task) and Table \ref{tab:c2m_iter} (for the Cap2Mol task).

From Table~\ref{tab:m2c_iter}, we can observe that the results without in-context reflection lead to sub-optimal performance as the teacher LLM could initially make mistakes or yield hallucinations.
However, the in-context reflected alignments are not necessarily better than zero-shot alignments, as evidenced by Table \ref{tab:c2m_iter}. 
As we utilize zero-shot alignments as context examples for self-reflection, the inaccuracies and hallucinations could be carried to the in-context reflected alignments, potentially harming the final performance. 
In this case, choosing between zero-shot alignments and in-context reflection alignments is imperative to ensure the quality of fine-grained alignments.

From an information theory perspective, our objective is to provide LLMs with more helpful information and less noise while rigorously preventing any disclosure of information about the target. Therefore, perplexity, an unsupervised metric, is an ideal criterion for the selection process. Higher perplexity scores suggest the presence of information that conflicts with the existing knowledge of LLMs, making it a reliable indicator for discerning the quality of the generated alignments. Theorem IV.1 also proves its effectiveness for reducing training cross-entropy.

In this work, we utilize Galactica-125M to calculate perplexity, which is particularly adept at chemical tasks and offers rapid computation. Those with the lower perplexity scores are selected as fine-grained alignments.
According to Table \ref{tab:m2c_iter} and \ref{tab:c2m_iter}, across both the Cap2Mol and Mol2Cap tasks, MolReFlect consistently demonstrates superior performance compared to those without in-context reflection or selection, although the performance gain is not that considerable, as the fine-grained alignments are obtained via sampling. 

Furthermore, we introduce semantic similarity as a supervised metric to assess the quality of alignments, as shown in Table \ref{tab:metric_m2c} and Table \ref{tab:metric_c2m}. The fine-grained alignments obtain the lowest perplexity, while maintaining a rather high semantic similarity towards the final target. This indicates that perplexity can serve as a good estimation of semantic similarity.

\noindent\textbf{RQ3: What is the necessity of adopting a teacher-student framework?}

We respond to this research question by ablating the student model and completing the tasks using only the teacher LLM (i.e., Llama-3-70B). Since the cost of fine-tuning the teacher LLM is beyond our capacity, we only test the performance of teacher LLM with prompt engineering to avoid modifications of their parameters. 
Various prompting strategies are implemented to enable the teacher LLM to undertake the molecule-caption translation tasks independently, including direct prompting, chain-of-thought prompting, few-shot prompting, and few-shot chain-of-thought prompting. Notably, in the chain-of-thought and few-shot chain-of-thought prompting, we utilize the fine-grained alignments produced by the teacher LLM itself as context information.
The results of these experiments are detailed in Table \ref{tab:m2c_pub} and \ref{tab:c2m_pub}. 

It can be observed that while Llama-3-70B is a powerful LLM, its performance under direct prompting is notably weak, as it is not previously trained on the ChEBI-20 nor related chemical corpora, ensuring that the information of the ChEBI-20 dataset is not leaked in its pre-training stage. 
In the Mol2Cap task, the chain-of-thought strategy enhances the performance by introducing fine-grained alignments. However, in the Cap2Mol task, the performance declines by 1.05\%, indicating that the teacher LLM struggles to filter out the inherent noise in the fine-grained alignments without explicit supervisory signals. Similarly, in the few-shot setting, the fine-grained alignments also fail to obtain a significant performance boost for the teacher LLM. In contrast, the student LLM proves to be indispensable and could benefit from the CoT-ICMT process by enabling a better understanding of molecule-text alignments and identifying the noises behind fine-grained alignments. As shown in Table \ref{tab:m2c_iter} and \ref{tab:c2m_iter}, the Instruction Tuning (i.e., w/o Context Examples) performance increases by 9.94\% and 14.22\% in the Mol2Cap and Cap2Mol tasks, respectively, compared to the naive-SFT.
This further underscores the necessity of discerning and mitigating noise within the fine-grained alignments, suggesting that LLMs must engage in fine-tuning to learn from the fine-grained alignments effectively.
Thus, the teacher-student framework proves to be indispensable. It enables the smaller student LLM to learn from the input distribution, discern noise in the content generated by the teacher, and absorb valuable information to inform the final generation process.

\section{Extensive Study}
Beyond the three research questions, we also conduct extensive studies to further prove and demonstrate the effectiveness of MolReFlect. Specifically, Sections VI.A and VI.B provide statistical view of the fine-grained alignments and the output distribution; Sections VI.C and VI.D proves the generalization capability of MolReFlect; Sections VI.E-G further demonstrate the robustness of MolReFlect; Sections VI.H conduct detailed case study and error analysis to show the explainability of our proposed method.

\subsection{Statistics of Fine-grained Alignments}
We evaluate the quality of fine-grained alignments with perplexity and an additional supervised metric, semantic similarity against the final target, calculated by sentencebert \cite{reimers2019sentence}. 
In the ideal scenario, fine-grained alignments provide all the necessary information to predict the final target, resulting in high semantic similarity.
As shown in Table \ref{tab:metric_m2c} and \ref{tab:metric_c2m}, as we select the fine-grained alignments by perplexity, the fine-grained alignments naturally inherit the lowest perplexity score. However, it is interesting to see that for the Mol2Cap task, the lower perplexity even indicates better semantic similarity to some extent, which is crucial for the generation of captions. Meanwhile, in the Cap2Mol task, selecting by lower perplexity also relieves the decreased semantic similarity of the in-context reflected alignments, further justifying the effectiveness of the selection strategy.

\begin{table}[htb]
    \centering
    \caption{Average semantic similarity and perplexity scores of different alignments and the original molecules in the training set for the Mol2Cap task (\textbf{Best}, \underline{Second Best}).}
    \resizebox{0.85\columnwidth}{!}{
    \begin{tabular}{c|c|c}
    \toprule
    Item & semantic similarity$\uparrow$ & perplexity $\downarrow$\\
    \midrule
    molecules & 0.2483 & 2.246 \\
    zero-shot alignments & 0.4983 & \underline{2.066}\\
    in-context reflected alignments & \underline{0.4985} & 2.070\\
    fine-grained alignments & \textbf{0.5029} & \textbf{1.995}\\
    \bottomrule
    \end{tabular}
    }
    \label{tab:metric_m2c}
\end{table}

\begin{table}[htb]
    \centering
    \caption{Average semantic similarity and perplexity scores of different alignments and the original molecule captions in the training set for the Cap2Mol task (\textbf{Best}, \underline{Second Best}).}
    \resizebox{0.85\columnwidth}{!}{
    \begin{tabular}{c|c|c}
    \toprule
    Item & semantic similarity$\uparrow$ & perplexity$\downarrow$ \\
    \midrule
    captions & 0.2483 & 2.758\\
    zero-shot alignments & \textbf{0.2721} & 2.426\\
    in-context reflected alignments & 0.2377 & \underline{2.351}\\
    fine-grained alignments & \underline{0.2524} & \textbf{2.230} \\
    \bottomrule
    \end{tabular}
    }
    \label{tab:metric_c2m}
\end{table}

\subsection{Output Distribution}
\begin{figure}[htbp]
    \centering
    \includegraphics[width=1.0\linewidth]{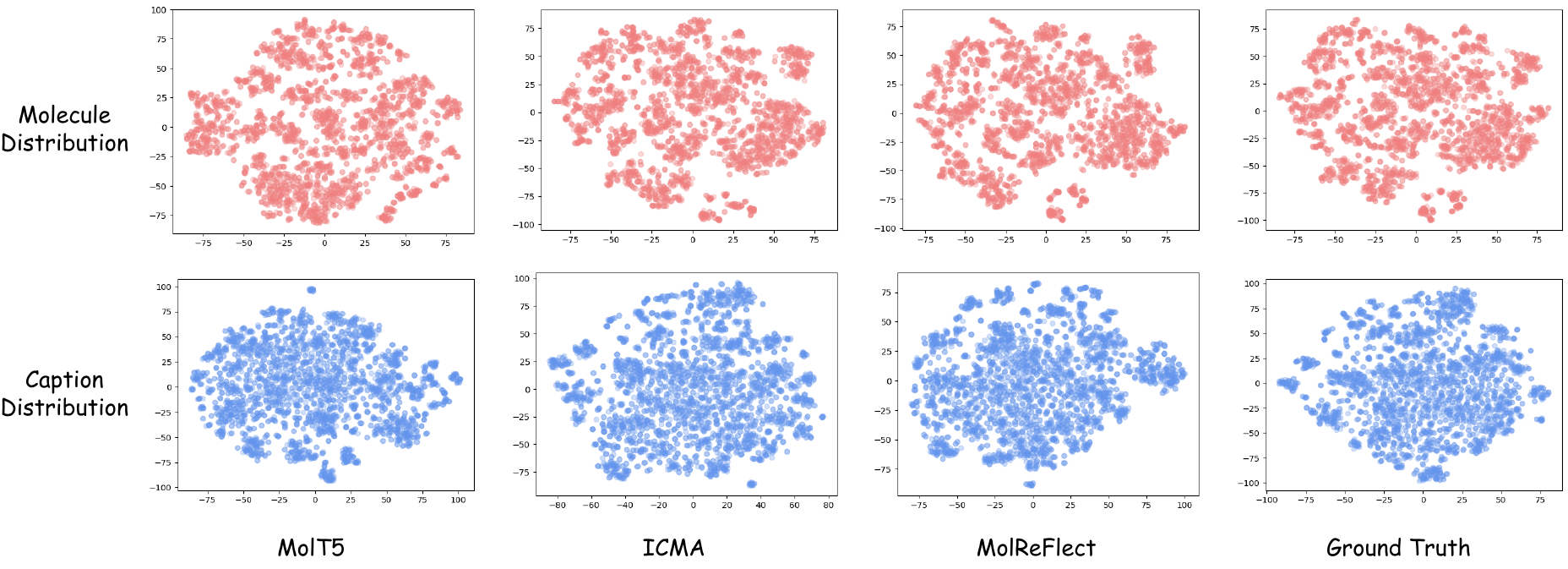}
    \caption{Embedding distributions of molecules and captions.}
    \label{fig:distribution}
\end{figure}

We also visualize the output distributions of different methods and the ground truth via sentencebert embeddings \cite{reimers2019sentence}, which are shown in Figure \ref{fig:distribution}. It is evident that the output distributions of MolT5 and ICMA are quite different: the caption distribution of MolT5 is more dense, while the caption distribution of ICMA is more sparse. However, MolReFlect generates a similar output distribution compared to the ground truth, better comprehending the mappings between molecules and texts.

\subsection{PubChem Performance}
To further illustrate the generalization performance of MolReFlect across different datasets, we conduct extensive experiments on another molecule-caption translation dataset, PubChem \cite{liu2023molca}. The results are shown in Table \ref{tab:pub_m2c} and Table \ref{tab:pub_c2m}. Here, we compare MolReFlect with the original Mistral-7B model for naive-supervised fine-tuning and ICMA \cite{li2024large}.

\begin{table}[htbp]
    \centering
    \caption{Mol2Cap Performance of MolReFlect on the PubChem dataset (\textbf{Best}, \underline{Second Best}). Here, Mistral-7B serves as the backbone LLM.}
    \resizebox{1.0\columnwidth}{!}{
    \begin{tabular}{c|c|c|c|c|c|c}
    \toprule
    Method & BLEU-2$\uparrow$ & BLEU-4$\uparrow$ & ROUGE-1$\uparrow$ & ROUGE-2$\uparrow$ & ROUGE-L$\uparrow$ & METEOR$\uparrow$ \\
    \midrule
    Mistral-7B & 0.361	& 0.288	& 0.471	& 0.325	& 0.419	& 0.421\\
    ICMA & \underline{0.369}	& \underline{0.297}	& \underline{0.482}&	\underline{0.342}	&\underline{0.433}	&\underline{0.431}\\
    \textbf{MolReFlect}& \textbf{0.414}	& \textbf{0.343}	& \textbf{0.511}&	\textbf{0.374}	&\textbf{0.458}	& \textbf{0.470}\\
    \bottomrule
    \end{tabular}
    }
    \label{tab:pub_m2c}
\end{table}

\begin{table}[htbp]
    \centering
    \caption{Cap2Mol Performance of MolReFlect on the PubChem dataset (\textbf{Best}, \underline{Second Best}). Here, Mistral-7B serves as the backbone LLM.}
    \resizebox{1.0\columnwidth}{!}{
    \begin{tabular}{c|c|c|c|c|c|c|c}
    \toprule
    Method & BLEU$\uparrow$ & EM$\uparrow$ & Levenshtein$\downarrow$ & MACCS FTS$\uparrow$ & RDK FTS$\uparrow$ & Morgan FTS$\uparrow$ & Validity$\uparrow$ \\
    \midrule
    Mistral-7B & 43.84	& 8.2 &	74.16	& 73.08	& 57.72 & 47.19 &	86.6\\
    ICMA & \underline{74.39}&	\underline{14.45}&	\underline{30.23}&	\underline{79.87}&	\underline{66.24}&	\underline{56.02}&	\underline{95.5}\\
    \textbf{MolReFlect}& \textbf{76.32}	& \textbf{17.15} &	\textbf{27.69} &	\textbf{80.6} &	\textbf{67.76} &	\textbf{57.65} &	\textbf{96.2} \\
    \bottomrule
    \end{tabular}
    }
    
    \label{tab:pub_c2m}
\end{table}

On both Mol2Cap and Cap2Mol tasks, MolReFlect still demonstrates the best performance on the PubChem dataset, significantly boosting the generation quality.
Meanwhile, the results also show a similar pattern to the ChEBI-20 dataset, proving the generalization capability of MolReFlect.

\subsection{Molecule Property Prediction}
Although our work is mainly focused on the molecule-caption translation task, we find its potential in molecule property prediction tasks. 
Here, we evaluate the MolReFlect performance on the BACE and BBBP tasks \cite{wu2018moleculenet}, two binary classification tasks. The results are listed in Table \ref{tab:mol_prop}. Here, we select Mistral-7B, ICMA(Mistral-7B), and MolReFlect (Mistral-7B) to ensure a fair comparison across the same backbone LLM.

\begin{table}[htb]
    \centering
    \caption{ROC-AUC (\%) scores of MolReFlect on the BACE and BBBP task from the MoleculeNet dataset \cite{wu2018moleculenet} (\textbf{Best}, \underline{Second Best}).}
    \begin{tabular}{c|c|c}
    \toprule
    Method & BACE & BBBP \\
    \midrule
    Mistral-7B	&0.4926	&0.4829 \\
    ICMA	&\underline{0.7995}	& \underline{0.6775} \\
    \textbf{MolReFlect}	& \textbf{0.8795}	&\textbf{0.8925}\\
    \bottomrule
    \end{tabular}
    \label{tab:mol_prop}
    \vskip -0.1in
\end{table}

The results show that MolReFlect achieves the best performance on the two molecule property prediction tasks, proving the potential of MolReFlect in generalizing to molecule property prediction tasks. 

\subsection{Comparing LLM-generated alignments with BRICS}
We also explore to use the BRICS algorithm \cite{landrum2013rdkit} to decompose molecules into substructures and replace the fine-grained alignments in our previous setting in Section V.A. Table \ref{tab:m2c_llama3} presents a quantitative comparison between the fine-grained alignments generated by LLMs and molecular substructures extracted by BRICS.

\begin{table}[htbp]
    \centering
    \caption{Comparing LLM-generated fine-grained alignments with BRICS substructure extraction (\textbf{Best}).}
    \resizebox{1.0\columnwidth}{!}{
    \begin{tabular}{c|c|c|c|c|c|c}
    \toprule
    Method & BLEU-2$\uparrow$ & BLEU-4$\uparrow$ & ROUGE-1$\uparrow$ & ROUGE-2$\uparrow$ & ROUGE-L$\uparrow$ & METEOR$\uparrow$ \\
    \midrule
    LLM& \textbf{0.676} & \textbf{0.608} & \textbf{0.703} & \textbf{0.571} & \textbf{0.644} & \textbf{0.680}\\
    BRICS & 0.671 & 0.603 & 0.699 & 0.567 & 0.640 & 0.676\\
    \bottomrule
    \end{tabular}
    }
    \label{tab:m2c_llama3}
\end{table}
The results demonstrate that LLM-generated fine-grained alignments consistently outperform the rule-based BRICS substructures across all the metrics. This suggests that LLMs can generate alignments that not only effectively decompose molecular structures but also hint the potential properties, which are similar to expert annotations.
The improvement, while incremental, validates the effectiveness of our teacher-student framework in learning to produce high-quality alignments without relying on predefined chemical rules.

\subsection{Model Agnosticism}
To verify the model agnosticism of MolReFlect, we conduct experiments on a different student LLM, Llama-3-8B-Instruct. We also remove the context examples and fine-grained alignments for ablation purposes.
The results are shown in Table \ref{tab:m2c_llama3} and \ref{tab:c2m_llama3}. We could observe similar trends in Llama-3-8B-Instruct compared to Mistral-7B: MolReFlect still achieves the best performance, and when removing context examples and fine-grained alignments, the performance all drops. Meanwhile, MolReFlect also empowers Llama-3-8B-Instruct to achieve SOTA performance on the ChEBI-20 dataset, demonstrating the model agnosticism of MolReFlect.

\begin{table}[htbp]
    \centering
    \caption{Mol2Cap Performance of MolReFlect when Llama-3-8B-Instruct serves as the student LLM (\textbf{Best}, \underline{Second Best}). We also compare the performance by removing the context examples and fine-grained alignments for ablation purposes.}
    \resizebox{1.0\columnwidth}{!}{
    \begin{tabular}{c|c|c|c|c|c|c}
    \toprule
    Method & BLEU-2$\uparrow$ & BLEU-4$\uparrow$ & ROUGE-1$\uparrow$ & ROUGE-2$\uparrow$ & ROUGE-L$\uparrow$ & METEOR$\uparrow$ \\
    \midrule
    \textbf{MolReFlect}& \textbf{0.672} & \textbf{0.605} & \textbf{0.703} & \textbf{0.571} & \textbf{0.644} & \textbf{0.678}\\
    w/o Examples & 0.617 & 0.540 & 0.661 & 0.515 & 0.598 & 0.622\\
    w/o Alignments & \underline{0.665} & \underline{0.595} & \underline{0.693} & \underline{0.559} & \underline{0.633} & \underline{0.669} \\
    \bottomrule
    \end{tabular}
    }
    \label{tab:m2c_llama3}
\end{table}

\begin{table}[htbp]
    \centering
    \caption{Cap2Mol Performance of MolReFlect when Llama-3-8B-Instruct serves as the student LLM (\textbf{Best}, \underline{Second Best}). We also compare the performance by removing the context examples and fine-grained alignments for ablation purposes.}
    \resizebox{1.0\columnwidth}{!}{
    \begin{tabular}{c|c|c|c|c|c|c|c}
    \toprule
    Method & BLEU$\uparrow$ & EM$\uparrow$ & Levenshtein$\downarrow$ & MACCS FTS$\uparrow$ & RDK FTS$\uparrow$ & Morgan FTS$\uparrow$ & Validity$\uparrow$ \\
    \midrule
    \textbf{MolReFlect}& \textbf{0.896} & \textbf{0.472} & \textbf{13.33} & \textbf{0.925}	& \textbf{0.846}	& \textbf{0.797}	&\textbf{0.979} \\
    w/o Examples& \underline{0.864} & 0.395  & \underline{16.13} & 0.904 & 0.815 & 0.754 & \underline{0.964}\\
    w/o Alignments& 0.851 & \underline{0.445}  & 19.27 & \underline{0.915} & \underline{0.836} & \underline{0.785} & 0.958\\
    \bottomrule
    \end{tabular}
    }
    
    \label{tab:c2m_llama3}
\end{table}

\begin{table*}[htbp]
    \centering
    \caption{Results of robustness probing test. The performance on the original test set is labelled as ``original" (\textbf{Best}, \underline{Second Best}).}
    \resizebox{0.95\linewidth}{!}{
    \begin{tabular}{c|c|c|c|c|c|c|c|c|c|c}
    \toprule
        \multirow{2}{*}{Probing Test} & \multicolumn{2}{c|}{MolT5-base} & \multicolumn{2}{c|}{Text+Chem T5-base} & \multicolumn{2}{c|}{MolT5-large} & \multicolumn{2}{c|}{Text+Chem T5-augm} & \multicolumn{2}{c}{MolReFlect} \\ \cline{2-11}
         & ROUGE-2 & METEOR & ROUGE-2 & METEOR & ROUGE-2 & METEOR & ROUGE-2 & METEOR & ROUGE-2 & METEOR \\
    \midrule
        original & 0.481 & 0.583 & 0.498 & 0.604 & 0.510 & 0.614 & \underline{0.543} & \underline{0.648} & \textbf{0.571} & \textbf{0.680} \\ 
        canonical & 0.315 & 0.450 & 0.381 & 0.515 & \underline{0.390} & \underline{0.532} & 0.377 & 0.514 & \textbf{0.416} & \textbf{0.543} \\ 
        hydrogen & 0.199 & 0.329 & 0.187 & 0.314 & 0.174 & 0.318 & \underline{0.201} & \underline{0.336} & \textbf{0.305} & \textbf{0.435} \\ 
        kekulization & 0.333 & 0.475 & \underline{0.413} & \underline{0.574} & 0.405 & 0.546 & 0.410 & 0.546 & \textbf{0.443} & \textbf{0.569} \\ 
        cycles & 0.417 & 0.540 & 0.483 & 0.600 & \textbf{0.566} & \underline{0.603} & 0.4575 & 0.581 & \underline{0.545} & \textbf{0.658} \\ 
    \bottomrule
    \end{tabular}
    }
    \vskip-0.1in
    \label{tab:rob}
\end{table*}

\subsection{Study of Model Robustness}
To verify the robustness of MolReFlect, we perform the probing test, following the work of ~\cite{ganeeva2024chemical} by transforming molecular SMILES into equivalent variants. Specifically, four different rules are applied:
\begin{itemize}
    \item \textbf{canonicalization}: Transforming a SMILES string into the RDKIT canonical SMILES string.
    \item \textbf{hydrogen}: Adding explicit hydrogen atoms into the SMILES string.
    \item \textbf{kekulization}: Transforming a SMILES string into the kekulized SMILES string. 
    \item \textbf{cycles}: Randomly replacing cycle numerical identifiers with other random numbers.
\end{itemize}

Here, we compare MolReFlect with the following baselines: MolT5-base, MolT5-large \cite{edwards-etal-2022-translation}, Text+Chem T5-base, and Text+Chem T5-augm \cite{christofidellis2023unifying}. The results are shown in Table \ref{tab:rob}.

The results show that although Text+Chem T5-augm achieves better original performance than MolT5-large, the augmentation makes it unrobust to the variance of molecule SMILES. However, MolReFlect not only achieves the highest score on the original test set but also shows the best robustness across the four SMILES variants.

\subsection{Case Studies}
\label{app:case}
\subsubsection{Fine-grained Alignment Cases}
\begin{figure}[htb]
    \centering
    \includegraphics[width=1.0\linewidth]{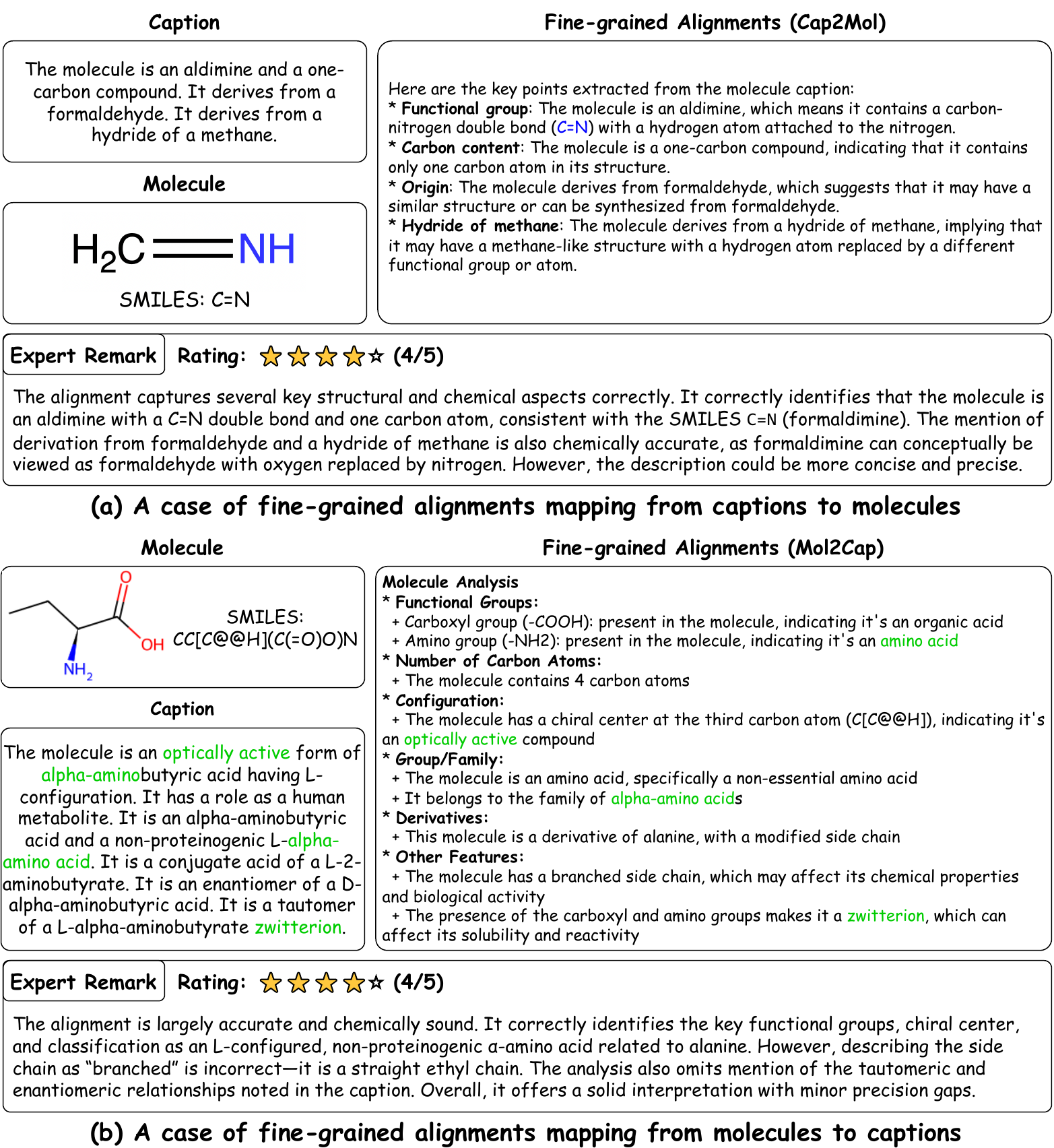}
    \caption{Cases of Fine-grained Alignments. We could observe that the molecule structure and characteristics have already been mentioned and aligned by the fine-grained alignments, which will surely benefit the final generations.}
    \label{fig:case_align}
\end{figure}

Figure \ref{fig:case_align} (a) and (b) depict the examples of fine-grained alignments. 
The larger teacher LLM can generate preliminary indications towards the final target and even directly figure out the molecular structure in fine-grained alignments. 
\subsubsection{Customized Cases}
\begin{figure}[htb]
    \centering
    \includegraphics[width=1.0\linewidth]{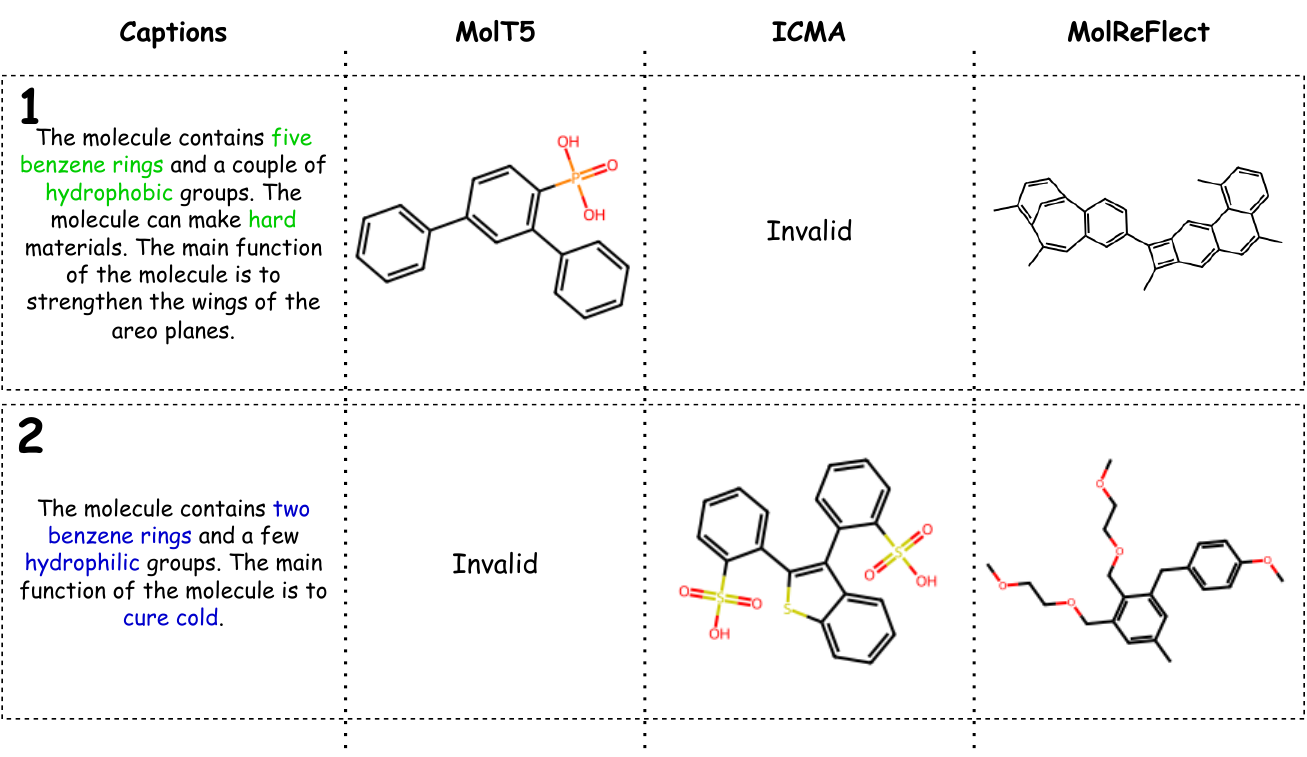}
    \caption{Cases of Customized Examples for the Cap2Mol task. We follow the customized examples in \cite{li2023empowering}. Obviously, MolReFlect generates correct molecules in general, matching the requirements mentioned in the customized cases, while MolT5 and ICMA fail to meet the requirements.}
    \label{fig:customize}
\end{figure}

We also include customized cases to highlight the superiority of MolReFlect compared with previous baselines. As shown in Figure \ref{fig:customize}, MolReFlect generates molecules that satisfy the requirements in the captions, while the other two methods fail.
\subsubsection{Error Analysis}
\begin{figure}[htb]
    \centering
    \includegraphics[width=1.0\linewidth]{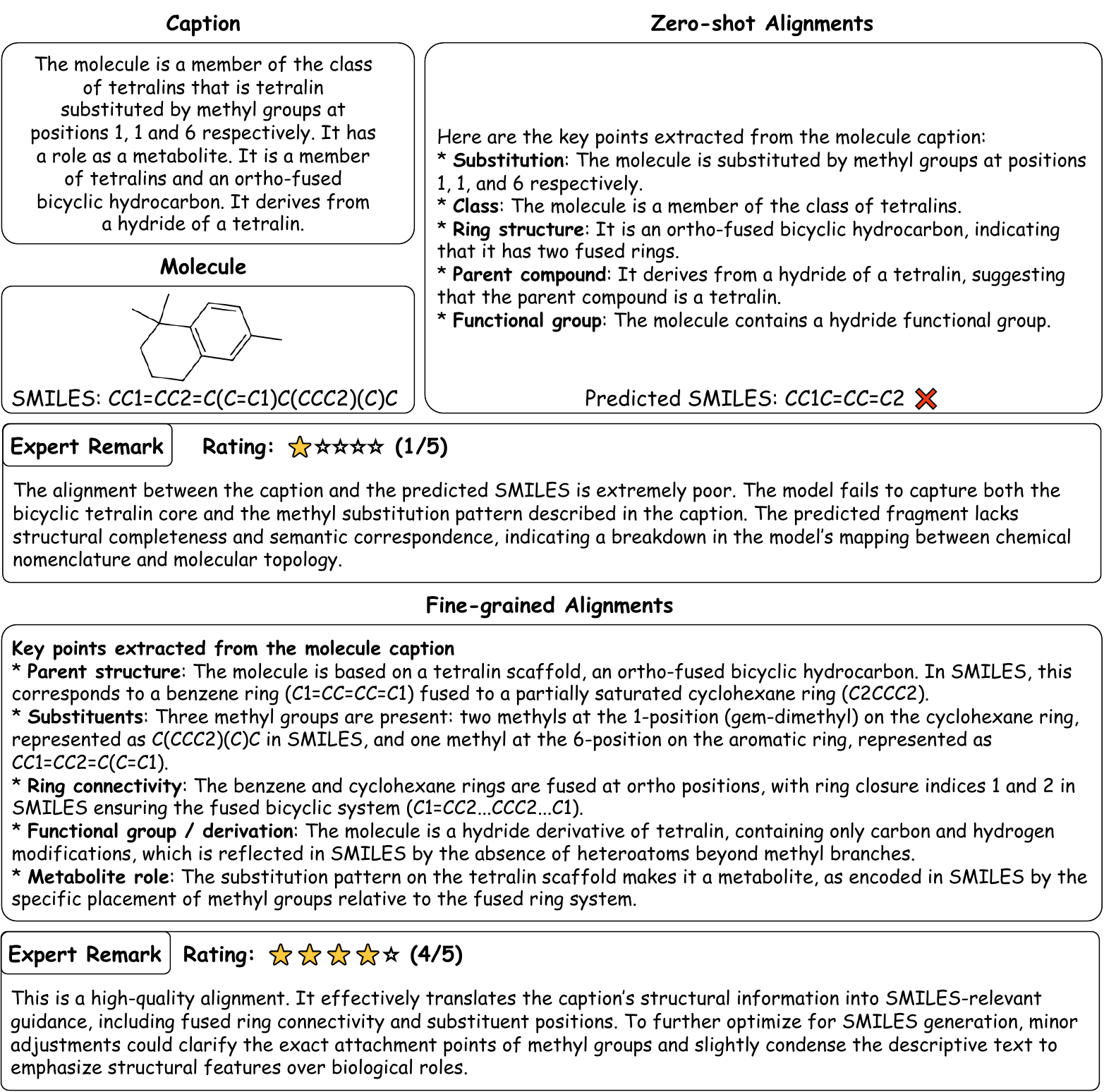}
    \caption{Error Analysis for the Cap2Mol task.}
    \label{fig:error}
\end{figure}

In Figure \ref{fig:error}, we conduct error analysis to further demonstrate the effectiveness of MolReFlect. It can be seen that the model initially fails because zero-shot alignments can not provide useful information for molecular structure prediction. However, the fine-grained alignments provide sufficient details for predicting the SMILES representation and successfully help the model to predict the correct molecule. 

\section{Conclusion}
\label{sec:conclusion}
In this study, we present MolReFlect, a novel teacher-student framework designed to refine the in-context alignments between molecular sub-structures and their corresponding textual descriptions. 
MolReFlect comprises three stages: Zero-shot Alignment Extraction, In-Context Selective Reflection, and Chain-of-Thought In-Context Molecule Tuning. 
Learning contextually from fine-grained alignments taught by the teacher LLM, the student LLM could better understand the detailed alignments between molecules and texts, enhancing the overall performance and contributing to a more explainable framework. 
Our experimental results reveal that MolReFlect outperforms all existing baselines on ChEBI-20. Additionally, we also substantiate the superior generalization, robustness, and explainability via extensive experiments and comprehensive case studies. We believe this work could inspire future works to focus on the granularity of molecule-text alignments in this promising field.

\section{Limitation}
Notably, MolReFlect has some limitations to be resolved in future works:
\begin{itemize}[itemsep=2pt,topsep=0pt,parsep=0pt]
    \item On one hand, the training cost of MolReFlect is larger than naive-supervised fine-tuning due to the expanded context length.
    \item On the other hand, due to the budget limitation and the worry of potential data leakage, we did not adopt the most advanced LLMs like GPT-4 and Claude-3.5 as the teacher LLM, although such fancy LLMs could bring greater improvements to the quality of fine-grained alignments.
\end{itemize}

\section*{Acknowledgments}
This research was supported by the National Natural Science Foundation of China (Grant No.: T2541073, 62372314). The experimental part of this work was supported by The Centre for Large AI Models (CLAIM) of The Hong Kong Polytechnic University.

\bibliographystyle{IEEEtran}

\bibliography{reference}
\vspace{-36pt}
{\begin{IEEEbiography}[{\includegraphics[width=1in,height=1.25in,clip,keepaspectratio]{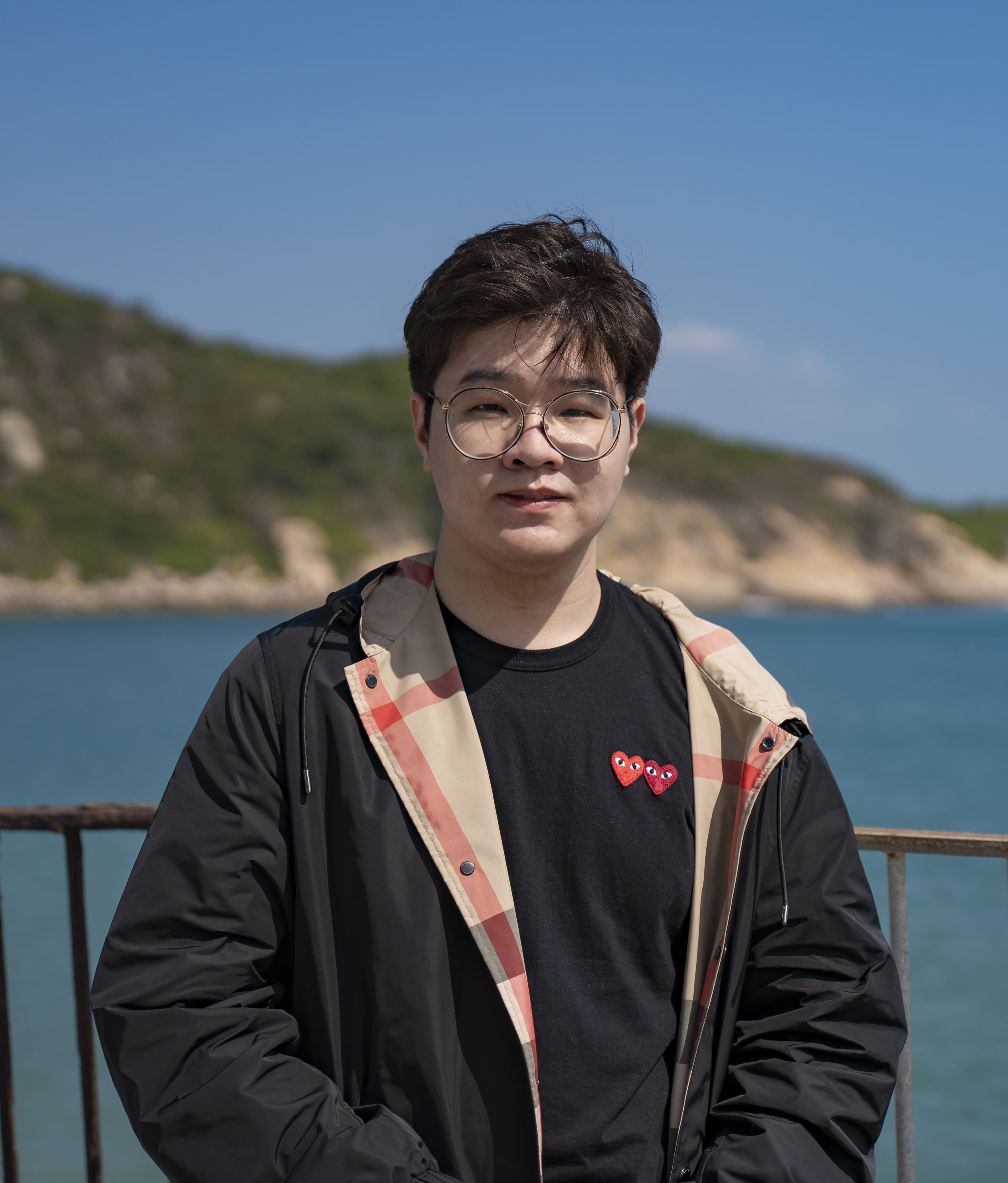}}]{Jiatong Li} is currently a PhD candidate of the Department of Computing (COMP), The Hong Kong Polytechnic University (funded by HKPFS). Before joining PolyU, he received his Master's degree in Information Technology (with Distinction) from the University of Melbourne, under the supervision of Dr. Lea Frermann. In 2021, he got his bachelor's degree in Information Security from Shanghai Jiao Tong University. His interest lies in Natural Language Processing, Drug Discovery, and Recommender Systems. He has published innovative works in top-tier conferences such as NAACL, IJCAI and ACL. For more information, please visit https://phenixace.github.io/.

\end{IEEEbiography}

\vspace{-28pt}
{\begin{IEEEbiography}[{\includegraphics[width=1in,height=1.25in,clip,keepaspectratio]{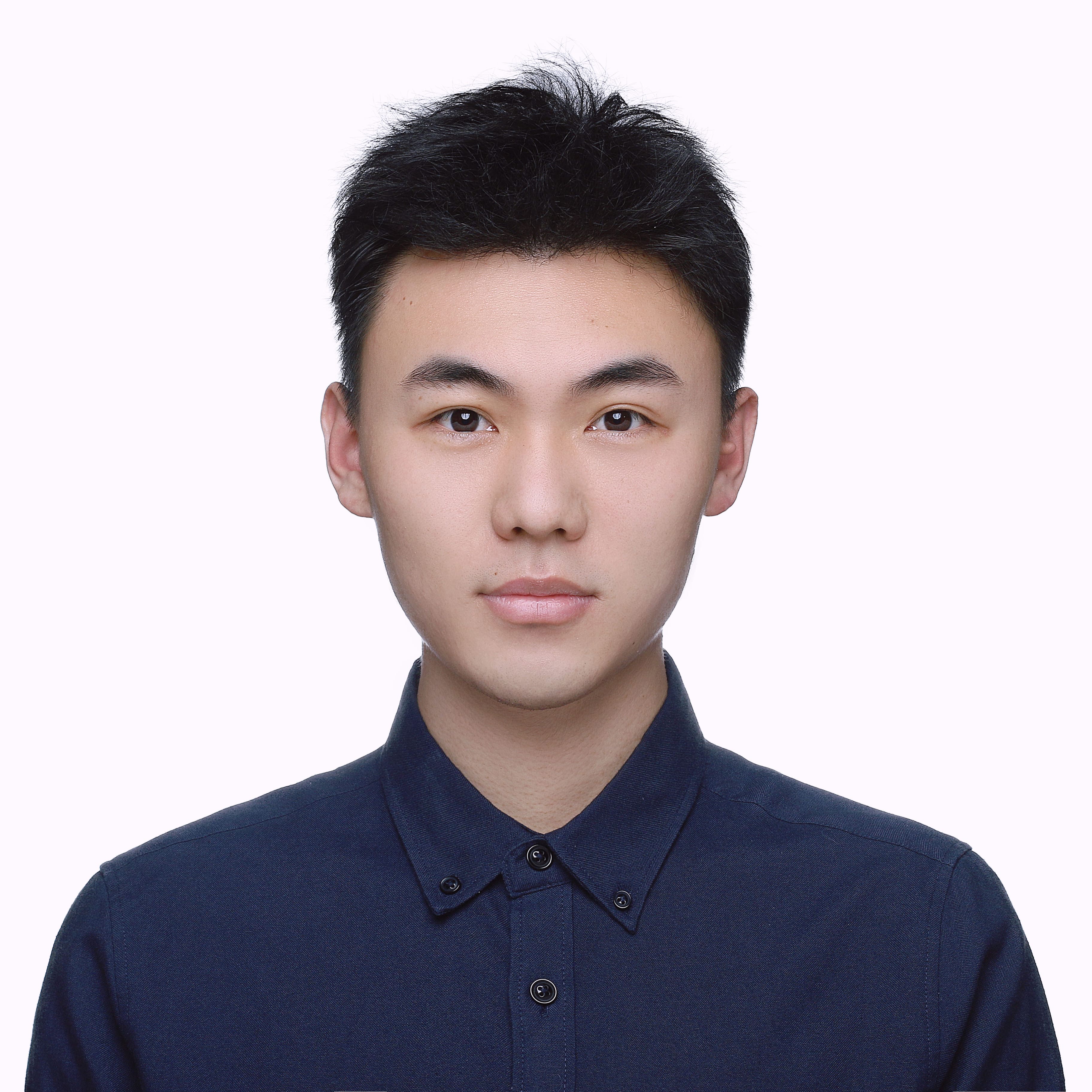}}]{Yunqing Liu} is currently a PhD candidate of the Department of Computing (COMP), Hong Kong Polytechnic University (PolyU), under the supervision of Dr. Wenqi Fan. Before joining the PolyU, he received his Master’s degree in Computer Science from the University of Edinburgh (M.Sc. in Computer Science), under the supervision of Dr. Elizabeth Polgreen. In 2020, he got his bachelor’s degrees from Wuhan University (B.Sc. in Chemistry and B.Eng. in Computer Science and Technology). His research interest includes Drug Discovery, Graph Neural Networks, and Natural Language Processing. He has published innovative works in top-tier conferences and journals such as IJCAI, EACL, EurJOC and Organic Letters. For more information, please visit https://liuyunqing.github.io/.

\end{IEEEbiography}

\vspace{-24pt}
{\begin{IEEEbiography}[{\includegraphics[width=1in,height=1.25in,clip,keepaspectratio]{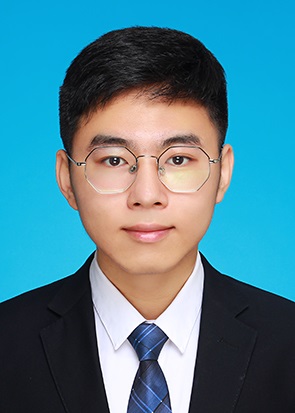}}]{Wei Liu} is currently a PhD student of the Department of Computer Science and Engineering, Shanghai Jiao Tong University, jointly trained with the Shanghai Artificial Intelligence Laboratory. In 2021, he got his bachelor's degree in Mechanical Engineering from Shanghai Jiao Tong University. His interest lies in AI-Driven Drug Design (AIDD) and Natural Language Processing. He has published works in conferences and journals such as ICLR and MBE.

\end{IEEEbiography}

\vspace{-24pt}
{\begin{IEEEbiography}[{\includegraphics[width=1in,height=1.25in,clip,keepaspectratio]{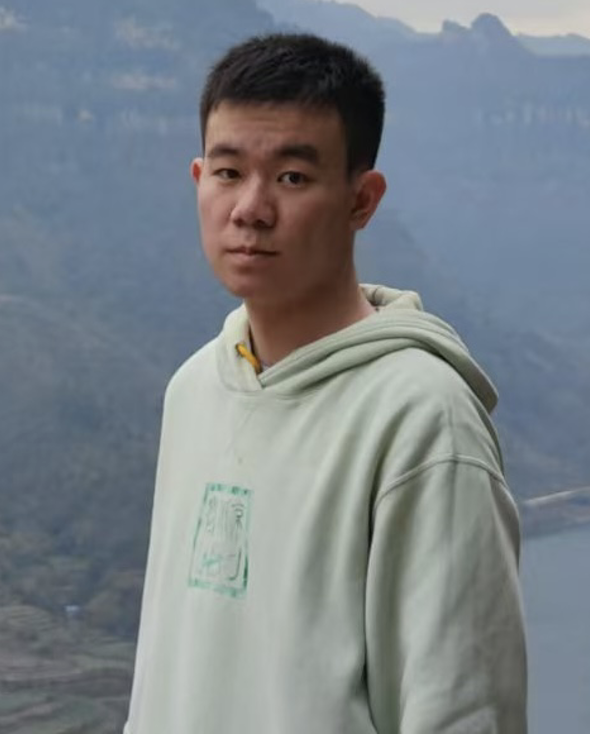}}]{Jingdi Lei} is currently a senior undergraduate student at Beijing Institute of Technology. He has been interning at the Shanghai Artificial Intelligence Laboratory. During his academic career, he consistently ranked first in his major for two consecutive years and was awarded the National Scholarship, the Dreaming Dongfeng Scholarship, and recognized as an Outstanding Student of Beijing Institute of Technology. His research focus at the laboratory has been on large-scale model mathematical reasoning, resulting in several publications at prestigious conferences including AAAI and NAACL.

\end{IEEEbiography}

\vspace{-24pt}
{\begin{IEEEbiography}[{\includegraphics[width=1in,height=1.25in,clip,keepaspectratio]{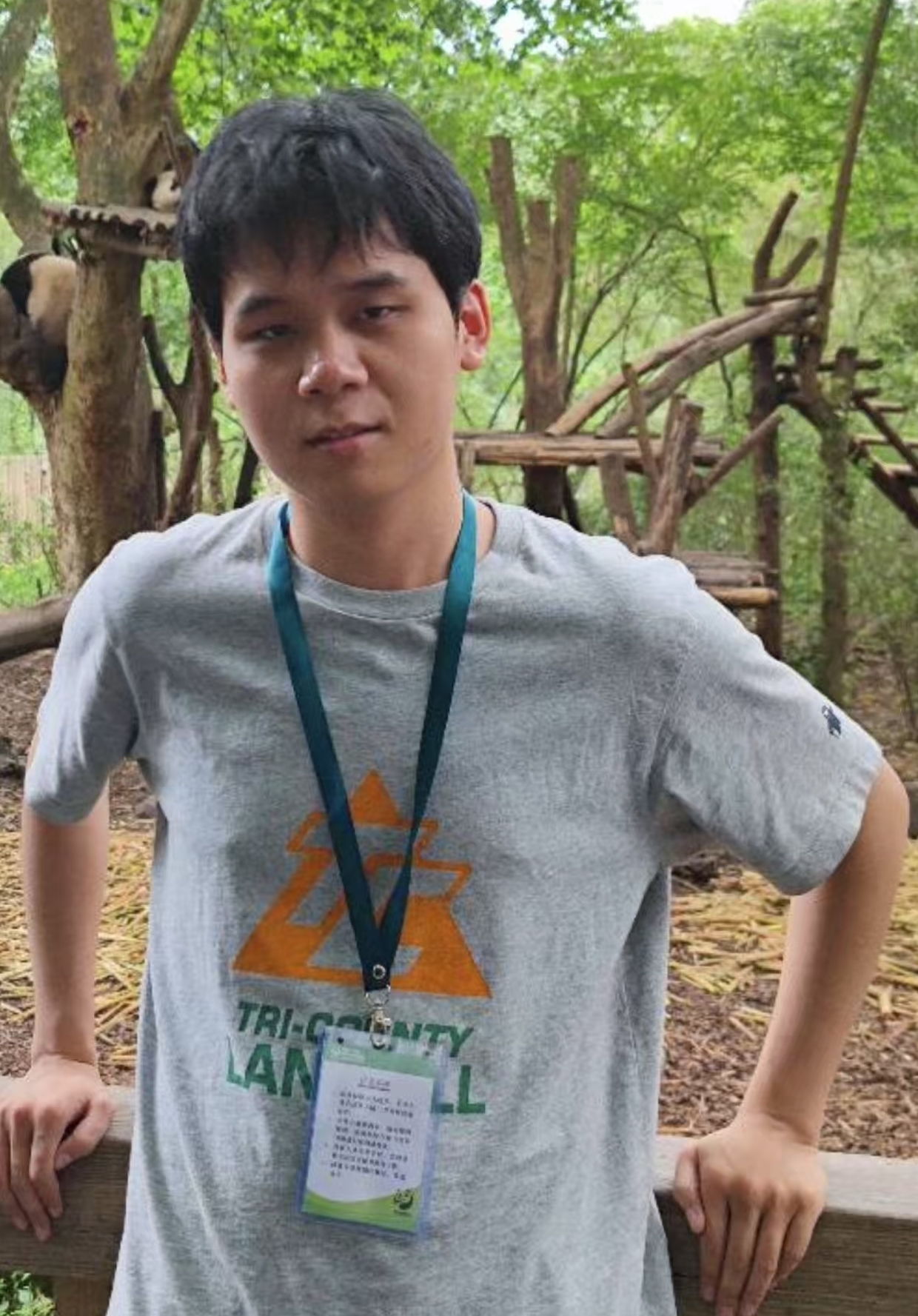}}]{Di Zhang} is currently a PhD candidate at Fudan University and the Shanghai Artificial Intelligence Laboratory, under the supervision of Professor Wanli Ouyang. Before joining Fudan University, he worked as a recommender system algorithm engineer at Alibaba Group. In 2022, he received his Master's degree in Computer Technology from the University of Science and Technology of China, under the supervision of Professors Xiaoping Chen and Bei Hua. In 2019, he obtained his Bachelor's degree in Water Supply and Drainage Science and Engineering from Hefei University of Technology. His research interests include AI4Chemistry, VLM, and LLM reasoning. He has published innovative works in top-tier conferences such as NAACL and AAAI, and served as a reviewer for ICLR and ICML. For more information, please contact di.zhang@ustc.edu. 

\end{IEEEbiography}



\vspace{-24pt}
{\begin{IEEEbiography}[{\includegraphics[width=1in,height=1.25in,clip,keepaspectratio]{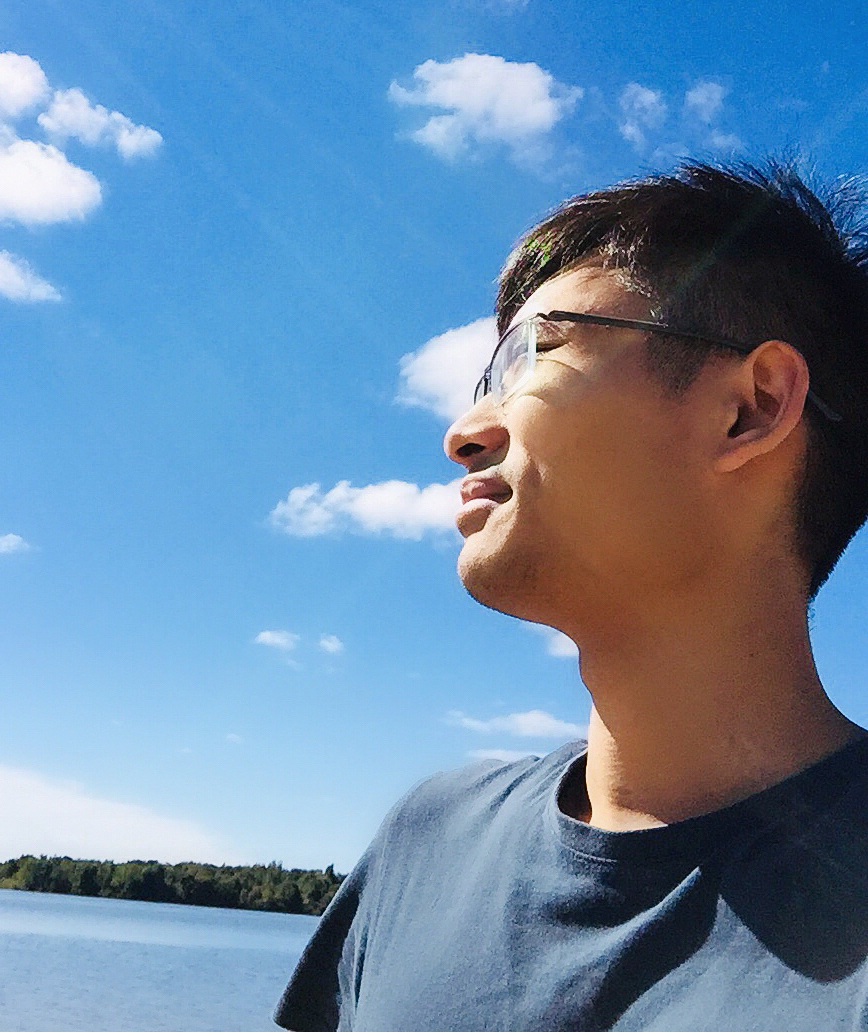}}]{Wenqi Fan} is a research assistant professor of the Department of Computing at The Hong Kong Polytechnic University (PolyU). He received his Ph.D. degree from the City University of Hong Kong (CityU) in 2020.
From 2018 to 2020, he was a visiting research scholar at Michigan State University (MSU). 
His research interests are in the broad areas of machine learning and data mining, with a particular focus on Recommender Systems, Graph Neural Networks, and Trustworthy Recommendations. He has published innovative papers in top-tier journals and conferences such as  TKDE, TIST, KDD, WWW, ICDE, NeurIPS, SIGIR, IJCAI, AAAI, RecSys, WSDM, etc. 
He serves as top-tier conference (senior) program committee members and session chairs (e.g., ICML, ICLR, NeurIPS, KDD, WWW, AAAI, IJCAI, WSDM, etc.), and journal reviewers (e.g., TKDE, TIST, TKDD, TOIS, TAI, etc.). 
More information about him can be found at https://wenqifan03.github.io.

\end{IEEEbiography}

\vspace{-24pt}
{\begin{IEEEbiography}[{\includegraphics[width=1in,height=1.25in,clip,keepaspectratio]{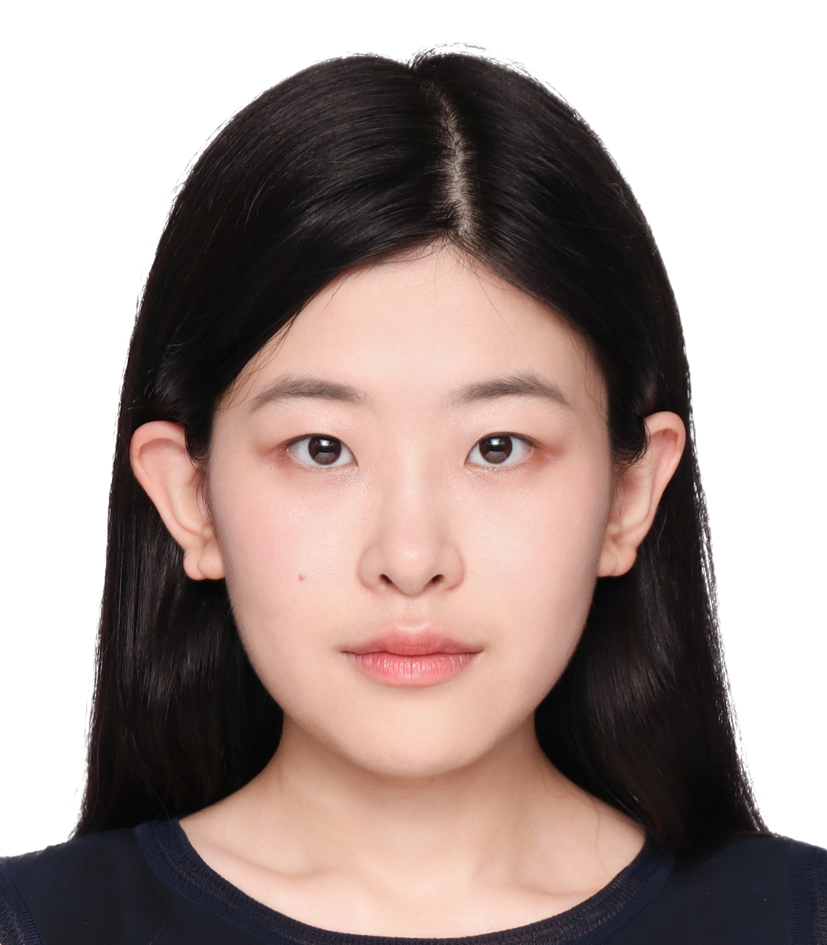}}] {Dongzhan Zhou} is currently a researcher at the AI for Science Center at Shanghai Artificial Intelligence Laboratory. Prior to this, she obtained her PhD and bachelor's degrees from the University of Sydney and Nanjing University, respectively. She has published several papers in top-tier journals and conferences in AI, chemistry, and materials science. Her research interests include foundation models (e.g., LLMs and multi-modal LLMs) and their applications in scientific discovery, especially in physical science.

\end{IEEEbiography}

\vspace{-24pt}
{\begin{IEEEbiography}[{\includegraphics[width=1in,height=1.25in,clip,keepaspectratio]{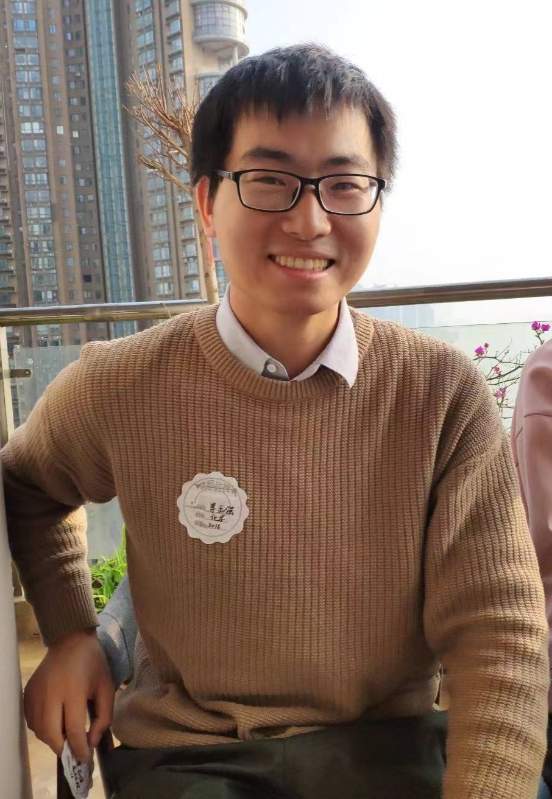}}] {Yuqiang Li} received a Bachelor of Science degree from Central South University in Changsha, China, and a Ph.D. in Chemistry from Wuhan University in Wuhan, China. He has previously held a lecturer position at Central South University. He is currently a junior researcher at the Shanghai AI Lab. His research interests encompass chemistry, machine learning, and large language models. He remains actively engaged in the scientific community, serving as a reviewer for journals such as Science Advance.

\end{IEEEbiography}

\vspace{-24pt}
\begin{IEEEbiography}
[{\includegraphics[width=1in,height=1.25in,clip,keepaspectratio]{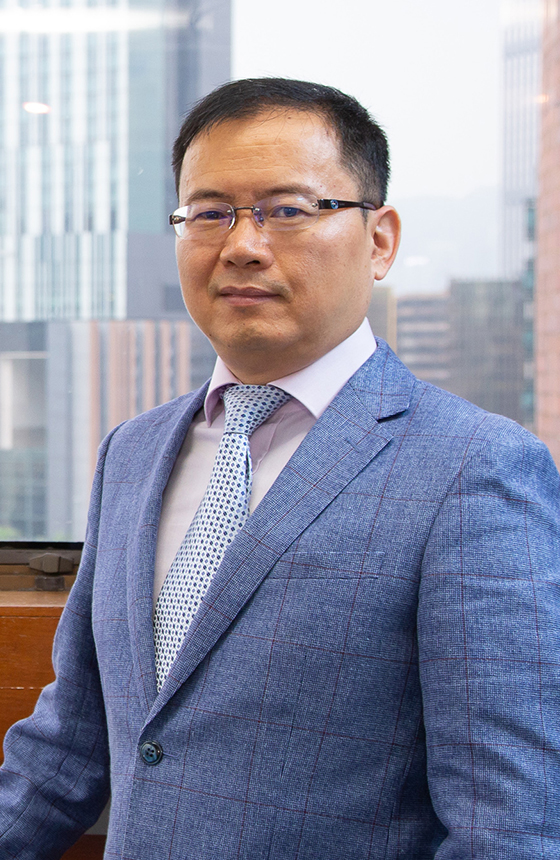}}]{Qing Li}
received the B.Eng. degree from Hunan University, Changsha, China, and the M.Sc. and Ph.D. degrees from the University of Southern California, Los Angeles, all in computer science.
He is currently a Chair Professor (Data Science) and the Head of the Department of Computing, the Hong Kong Polytechnic University. He is a Fellow of IEEE and IET, a member of ACM SIGMOD and IEEE Technical Committee on Data Engineering. 
His research interests include object modeling, multimedia databases, social media, and recommender systems. 
He has been actively involved in the research community by serving as an associate editor and reviewer for technical journals, and as an organizer/co-organizer of numerous international conferences. 
He is the chairperson of the Hong Kong Web Society, and also served/is serving as an executive committee (EXCO) member of IEEE-Hong Kong Computer Chapter and ACM Hong Kong Chapter. In addition, he serves as a councilor of the Database Society of Chinese Computer Federation (CCF), a member of the Big Data Expert Committee of CCF, and is a Steering Committee member of DASFAA, ER, ICWL, UMEDIA, and WISE Society. 
\end{IEEEbiography}
} 

\end{document}